\documentclass{article}

 \usepackage[eandd, final]{neurips_2026}

\usepackage[utf8]{inputenc} 
\usepackage[T1]{fontenc}    
\usepackage{hyperref}       
\usepackage{url}            
\usepackage{booktabs}       
\usepackage{amsfonts}       
\usepackage{nicefrac}       
\usepackage{microtype}      
\usepackage{xcolor}         
\usepackage{wrapfig}
\usepackage{multirow}
\usepackage{graphicx}
\usepackage{caption}
\usepackage{tcolorbox}
\usepackage{bbm}
\usepackage{fvextra}
\usepackage[table]{xcolor}
\usepackage{adjustbox}

\title{VIGOR: Benchmarking Visual Rationale Correctness in Multimodal Large Language Models}

\author{%
  Yanlin Chen \\
  University of Virginia\\
  \And
  Tang Li \\
  University of Delaware \\
  \And
  Xi Peng \\
  University of Virginia \\
}

\begin{document}

\maketitle

\begin{abstract}
Multimodal large language models (MLLMs) are typically evaluated by whether they produce correct answers, but correctness alone does not reveal whether their outputs are supported by the right visual evidence. We introduce VIGOR, a benchmark for evaluating \textbf{VI}sually \textbf{G}r\textbf{O}unded \textbf{R}ationales in MLLMs. VIGOR represents visual rationales as explicit concepts paired with pixel-level groundings, covering multiple abstraction levels including low-level visual properties, intermediate structures, objects, parts and high-level motion, relations. To construct VIGOR at scale, we develop an automatic annotation framework that reuses existing dense supervision when available and generates missing concept groundings through proposal-based annotation followed by conservative multimodal verification. Using VIGOR, we evaluate whether MLLMs ``look at what they say'' by comparing model-derived visual evidence for each expressed concept against annotated ground-truth rationales. Our results show that current MLLMs exhibit limited visual rationale correctness and reveal distinct grounding failures across concept levels. VIGOR provides a benchmark and scalable data construction pipeline for studying whether MLLM outputs are supported by appropriate visual evidence. Code is available at \url{https://github.com/deep-real/VIGOR}
\end{abstract}

\section{Introduction}

\begin{wrapfigure}{r}{0.42\textwidth}
\vspace{-10pt}
\centering
\includegraphics[width=\linewidth]{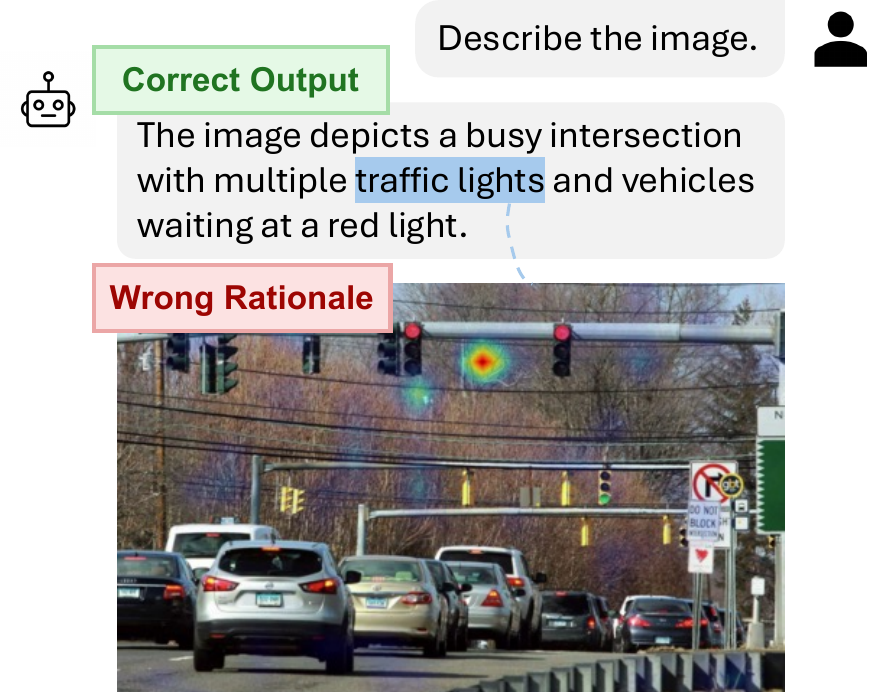}
\caption{
Illustration of the \textbf{double correctness} problem. Although the model produces a correct description by mentioning ``traffic lights,'' its visual rationale is misaligned with the actual traffic light regions, showing the gap between \textit{output correctness} and \textit{rationale correctness}.
}
\vspace{-25pt}
\label{fig:overview}
\end{wrapfigure}
Trustworthy visual systems should be correct not only in their final outputs, but also in the visual evidence that supports those outputs. We define a model's \emph{rationale} as its internal visual grounding for an output: the image evidence used by the model during prediction, \textit{i.e.}, where the model ``looks at'' when producing its answer. Unlike output-level grounding~\cite{lai2024lisa, zhang2024llava, ren2024pixellm}, which links generated words to image regions after prediction, rationale correctness evaluates whether the model itself relied on the appropriate visual evidence. A model may correctly classify an image, answer a visual question, or generate a plausible description while relying on the wrong object, an irrelevant background region, or a spurious contextual cue~\cite{ribeiro2016should,goyal2017making,geirhos2020shortcut}. Such behavior creates a gap between \emph{output correctness} and \emph{rationale correctness}: the prediction is right, but the underlying visual basis is wrong. We refer to this requirement as the \textbf{double correctness} problem (Figure~\ref{fig:overview}). A reliable visual system should satisfy both conditions: it should produce correct outputs based on appropriate image evidence.

This problem is especially important for multimodal large language models (MLLMs). Recent advances in MLLMs have produced systems that can answer image-based questions, describe complex scenes, and generate fluent explanations about visual content~\cite{li2023blip2,liu2023visual,zhu2023minigpt4,bai2025qwen25vl}. However, the fluency of these models can obscure whether their generated concepts are actually supported by the image. For example, in radiology, an MLLM may produce a plausible chest-X-ray report that identifies pneumonia, yet localize the evidence to the wrong lung region or rely on acquisition artifacts rather than the pathological opacity itself~\cite{mahmood2025phrase,degrave2021ai}. These failures raise a central question for trustworthy multimodal understanding: \emph{do MLLMs visually ground the concepts they express?}

Existing MLLM benchmarks are not designed to answer this question. Most evaluate output correctness and general multimodal capability
\citep{fu2026mme, liu2024mmbench, li2024seed, yue2024mmmu},
hallucination
\citep{li2023pope, guan2024hallusionbench},
or instruction following and open-ended response quality
\citep{liu2023visual, yu2023mmvet}, but they do not specify the \textit{visual evidence} that should support each generated concept. Grounding datasets provide useful annotations, yet their scope is typically limited to objects, regions, or referring expressions~\cite{yu2016modeling,krishna2017visual,kazemzadeh2014referitgame,rasheed2024glamm}. They offer little pixel-level annotations for broader visual concepts such as color, edge, texture, shape, motion, and relation. As a result, current evaluations can measure whether an MLLM says the right thing, but they provide limited insight into whether the model relies on the right visual evidence when producing that content. Without benchmarks that directly evaluate visual rationales, models that produce correct answers for the wrong visual reasons may remain indistinguishable from models that are genuinely grounded, limiting our ability to diagnose models' failures and improve their trustworthiness.

To address this gap, we introduce VIGOR, a benchmark for evaluating \textbf{Vi}sually \textbf{G}r\textbf{O}unded \textbf{R}ationales in MLLMs. VIGOR represents visual rationales as explicit concepts paired with pixel-level evidence in the image. Its concept space spans multiple abstraction levels, including low-level visual properties, intermediate structures, objects, parts and high-level motion, relations. This design enables evaluation beyond object-centric grounding: instead of asking only whether a model can localize a noun phrase, VIGOR asks whether the model's visual evidence aligns with the full range of concepts it describes.

Constructing such a benchmark is challenging because existing datasets provide only partial supervision across the visual concept hierarchy. Object and part masks can often be reused from existing resources, but many other concept types lack direct pixel-level annotations. We therefore develop an automatic annotation framework that combines heterogeneous annotation sources with \textit{proposal-and-verification} based annotation. Existing masks are reused when available, while missing concept groundings are generated through complementary proposal tools and filtered by conservative multimodal verification. This allows VIGOR to scale beyond the narrow coverage of existing grounding datasets while maintaining reliable pixel-level rationales.

Using VIGOR, we benchmark whether MLLMs \textit{``look at what they say''} by comparing model-derived visual evidence for each concept against its annotated grounding. Our experiments reveal that current MLLMs still have limited visual rationale correctness, with the best model only achieving $31.2\%$ AP and $38.7\%$ Pointing Accuracy (Sec.~\ref{sec:main_results}). We also find that different model families, sizes and model layers exhibit distinct failure modes (Sec.~\ref{sec:layerwise_analysis} - \ref{sec:error_analysis}). These results suggest that visually grounded rationale evaluation provides a complementary perspective on MLLM trustworthiness compared to standard hallucination benchmarks.

Our contributions are fourfold. \textbf{First}, we formulate visually grounded rationale evaluation as a benchmark problem for studying double correctness in MLLMs, focusing on whether models ground the concepts they express in the correct visual evidence. \textbf{Second}, we introduce VIGOR, a dataset with pixel-level annotations for visual concepts across multiple abstraction levels, extending beyond the object- and part-centric scope of existing grounding resources. \textbf{Third}, we propose an automatic annotation framework that makes broad-coverage rationale annotation feasible by combining existing supervision, generated proposals, and multimodal verification. \textbf{Finally}, we conduct a large-scale evaluation across diverse MLLM families and model sizes, revealing that current MLLMs still struggle to localize the visual evidence underlying the concepts they express, and providing diagnostic insights that can guide the development of more visually grounded and trustworthy MLLMs.
\section{Related Work}

\noindent \textbf{Visual grounding and dense visual annotations.}
Existing datasets provide dense supervision for objects, regions, parts, attributes, and referring expressions. COCO~\cite{lin2014microsoft} and COCO-Stuff~\cite{caesar2018coco} annotate objects and stuff regions, LVIS~\cite{gupta2019lvis} expands instance segmentation to long-tailed objects, Visual Genome~\cite{krishna2017visual} covers objects, attributes, and relations, RefCOCO~\cite{yu2016modeling} variants and PhraseCut~\cite{wu2020phrasecut} ground language phrases, and PACO~
\cite{ramanathan2023paco} adds part masks and attributes. While these resources are essential for grounding, they are usually specialized to specific concept types. VIGOR instead covers a broader hierarchy of visual concepts and provides pixel-level evidence for each concept, enabling systematic evaluation of visually grounded rationales.

\noindent \textbf{Visual rationales and explanation datasets.}
Datasets such as VQA-X~\cite{park2018multimodal}, VQA-HAT~\cite{das2017human}, and e-SNLI-VE~\cite{do2020snli} study visual explanations through textual rationales, human attention maps, or task-specific evidence. These benchmarks evaluate explanation plausibility or human-aligned attention in downstream tasks, but they rarely specify spatial evidence for individual visual concept. VIGOR treats rationales as concept-level visual evidence: for each annotated concept, it specifies the supporting image region, allowing direct evaluation of whether models ground the concepts they process.

\noindent \textbf{MLLM evaluation and hallucination benchmarks.}
MLLM benchmarks such as MME~\cite{fu2026mme}, MMBench~\cite{liu2024mmbench}, SEED-Bench~\cite{li2024seed}, MMMU~\cite{yue2024mmmu}, CHAIR~\cite{rohrbach2018object}, POPE~\cite{li2023pope}, and HallusionBench~\cite{guan2024hallusionbench} evaluate perception, reasoning, instruction following, and hallucination at the output level. They measure whether models produce correct responses, but generally do not inspect the visual evidence underlying those responses. VIGOR complements these evaluations by measuring rationale correctness: whether the visual tokens associated with a concept align with the correct image regions.
\section{The VIGOR Benchmark}

\begin{figure}
  \centering
  \includegraphics[width=\textwidth]{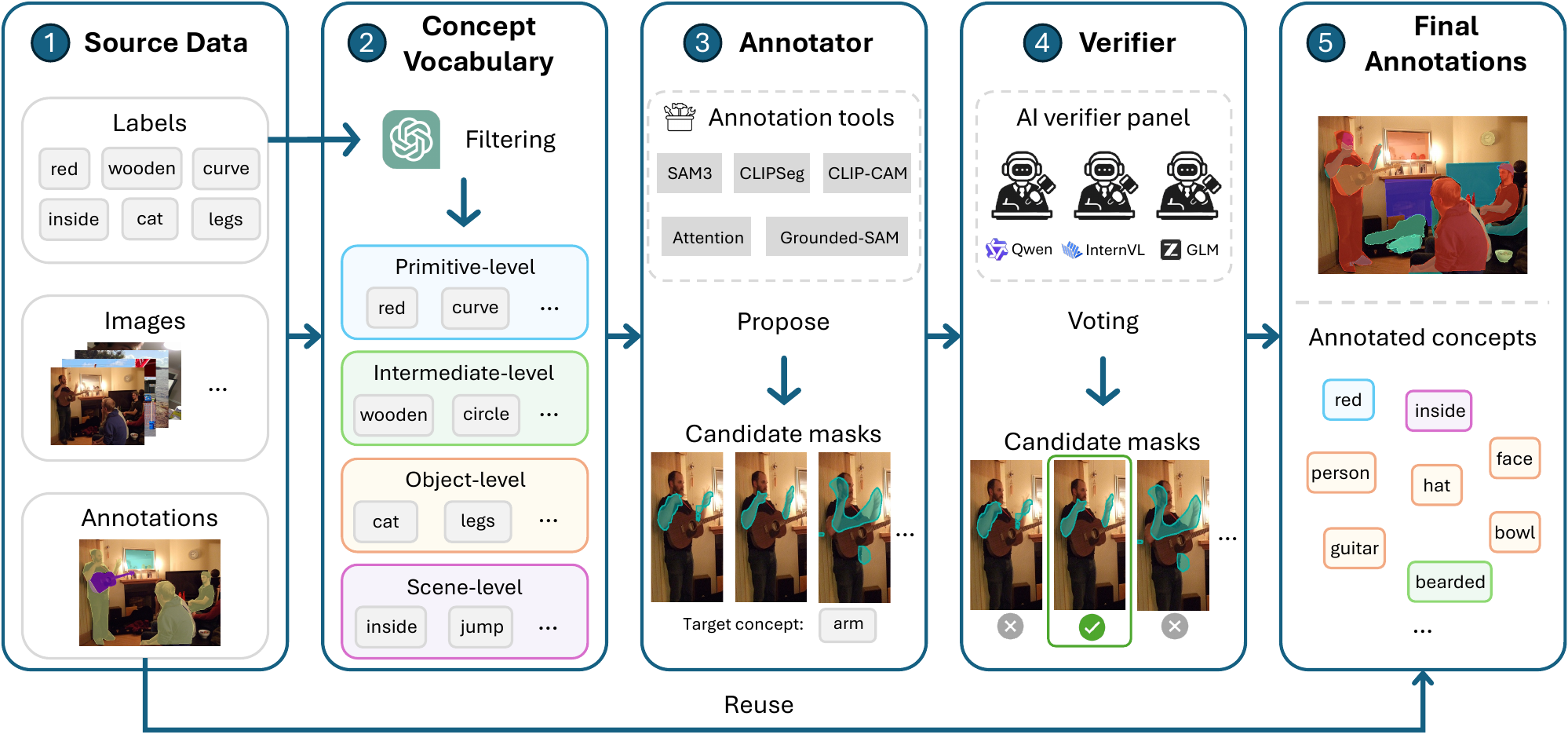}
\caption{
Overview of the VIGOR annotation pipeline. VIGOR builds a hierarchical concept vocabulary from heterogeneous source data, proposes masks with multiple annotation tools when source masks are unavailable, verifies candidate masks through MLLM voting, and combines verified masks with reusable source annotations to produce pixel-level visual rationales.
}
\label{fig:pipeline}
\end{figure}
\subsection{Overview of VIGOR}

We introduce VIGOR, a benchmark for evaluating visually grounded rationales. Each image in VIGOR is paired with visual concepts spanning multiple abstraction levels, together with pixel-level annotations that identify the image evidence supporting each concept. Importantly, our goal is not to provide perfect object segmentation or exhaustively precise pixel-level masks, but rather to provide visual evidence for each concept. Such evidence may be coarse as long as it reliably captures the image regions that support the concept. These annotations are model- and attribution-agnostic: they specify the evidence a correct rationale should cover, against which attention maps, gradient-based attributions, or future causal explanations can all be scored. The central challenge is that existing datasets provide only partial supervision over the visual concept hierarchy: object- and part-level concepts are relatively well annotated, while attributes, motions, relations, and other higher- or lower-level concepts rarely have dense visual groundings. This raises a key question: \textit{how can we construct grounded rationales at scale for a broad concept space?}

To address this challenge, we develop an automatic annotation framework (illustrated by Figure~\ref{fig:pipeline}) that integrates heterogeneous annotation sources into a unified pipeline. Existing annotations are reused whenever they already capture the target concept, while concepts outside standard object- and part-level annotation are grounded through automatically generated proposals followed by conservative verification. By combining source reuse, concept completion, mask proposal, and multimodal verification, VIGOR extends grounded annotation beyond the limited scope of existing datasets and enables large-scale rationale annotation across the visual concept hierarchy.

\subsection{Automatic Annotation Framework}

\noindent \textbf{Concept vocabulary.}
Grounded rationale annotation requires a vocabulary that defines \emph{what} can be annotated. We organize concepts into four abstraction levels: \emph{primitive} concepts such as colors, edges, and curves; \emph{intermediate} concepts such as textures, materials, and shapes; \emph{object} concepts covering objects and parts; and \emph{scene} concepts covering actions and spatial relations. We collect candidate concepts from existing datasets, including DTD~\cite{cimpoi2014describing}, VAW~\cite{pham2021learning}, Visual Genome~\cite{krishna2017visual}, PACO~\cite{ramanathan2023paco}, COCO~\cite{lin2014microsoft}, ADE20K~\cite{zhou2017scene}, LVIS~\cite{gupta2019lvis}, Open Images~\cite{kuznetsova2020open}, AVA~\cite{gu2018ava}, Kinetics~\cite{kay2017kinetics}, and GQA~\cite{hudson2019gqa}, and further expand long-tail coverage using COCO and Visual Genome captions. This yields an initial pool of 21,352 candidate concepts.

The raw concept pool is broad but noisy, containing redundant, invalid, and overly fine- or coarse-grained entries. We therefore filter the candidates using multiple LLMs according to four criteria: (1) \textit{linguistic validity}, removing malformed entries such as ``of the'' or ``standing in''; (2) \textit{visual groundability}, removing non-visual concepts such as ``important'' or ``happy''; (3) \textit{atomicity}, removing compound concepts such as ``red striped shirt''; and (4) \textit{appropriate semantic granularity}, removing concepts that are too broad or too specific, such as ``object'' or ``2012 Honda Civic''. For each candidate concept, the providers assign a concept-level label and decide whether it satisfies these criteria. A concept is retained only if at least two LLMs choose to keep it. Finally, we conduct a cleaning pass using LLM to remove residual duplicates, lexical variants, and category inconsistencies. The resulting VIGOR vocabulary contains 3,011 concepts across the four concept levels. For choices of LLMs and prompts used for concept filtering and cleaning, see \S Appendix~\ref{app:concept_vocab}.

\noindent \textbf{Data sourcing.}
Given the vocabulary, we construct a concept-complete annotation set for each image. Because no single dataset covers the full VIGOR concept hierarchy, we first consolidate complementary annotations from COCO, COCO-Stuff, LVIS, PACO-LVIS, and Visual Genome, and normalize their labels into the unified VIGOR concept space. This provides an initial concept inventory for each image, but remains incomplete because many visible concepts are not annotated in the source datasets. To improve coverage, we use a MLLM to inspect each image and identify additional visible concepts from the final vocabulary that are missing from the normalized source annotations. We then select the 1,000 images with the largest number of concepts for annotation. For more details on missing concept identification, see \S Appendix~\ref{app:data_sourcing}.

\noindent \textbf{Annotator.}
The annotator converts each completed image-level concept set into pixel-level grounding proposals. Its strategy depends on the annotation available for each concept. Concepts with existing source masks are directly converted into binary masks, while concepts with only bounding-box annotations are processed with box-conditioned mask generation to refine coarse localization into pixel-level support. For concepts without direct annotations, we use a diverse set of complementary annotators, including CLIP CAM \citep{chefer2021transformer}, MLLM attention-map ensembles, SAM3~\cite{carion2025sam}, Grounded-SAM \citep{ren2024grounded}, and CLIPSeg \citep{luddecke2022image}. These tools provide different forms of visual evidence: CLIP CAM capture image regions semantically aligned with the target concept, attention ensembles exploit internal localization signals from MLLMs, SAM3 and Grounded-SAM produce high-quality region proposals and text-conditioned segmentations, and CLIPSeg provides open-vocabulary segmentation masks. Using multiple annotators is important because no single tool reliably grounds all concept levels: low-level attributes, object parts, full objects, actions, and spatial relations often require different localization cues. The resulting proposal set therefore improves coverage and robustness by allowing later verification to select reliable evidence while rejecting noisy or inconsistent masks. For relation concepts, whose visual evidence often lies in the configuration between entities rather than in a single object region, the annotator may additionally construct relation masks from subject and object boxes to cover the supporting interaction area. Full details on annotator are provided in \S Appendix~\ref{app:annotator}.



\begin{figure}[t]
    \centering
    
    \begin{minipage}[t]{0.37\textwidth}
        \centering
        \vspace{0pt}
        \small
        \captionof{table}{Key statistics of VIGOR.}
        \resizebox{\linewidth}{!}{
        \begin{tabular}{l r}
            \toprule
            \textbf{Statistic} & \textbf{Number} \\
            \midrule
            \# images & 1000 \\
            \# unique concepts & 3011 \\
            Avg masks per image & 31.4 \\
            \midrule
            \# masks & \\
            \hspace{0.5em}* Primitive level & 1993 \\
            \hspace{0.5em}* Intermediate level & 1214 \\
            \hspace{0.5em}* Object level & 23863 \\
            \hspace{0.5em}* Scene level & 4363 \\
            \midrule
            Sourced masks & 18038 \\
            Annotated masks & 13395 \\
            \bottomrule
        \end{tabular}
        }
        \label{tab:dataset_statistics}
    \end{minipage}
    \hspace{20pt}
    \begin{minipage}[t]{0.5\textwidth}
        \centering
        \vspace{0pt}
        \includegraphics[width=\linewidth]{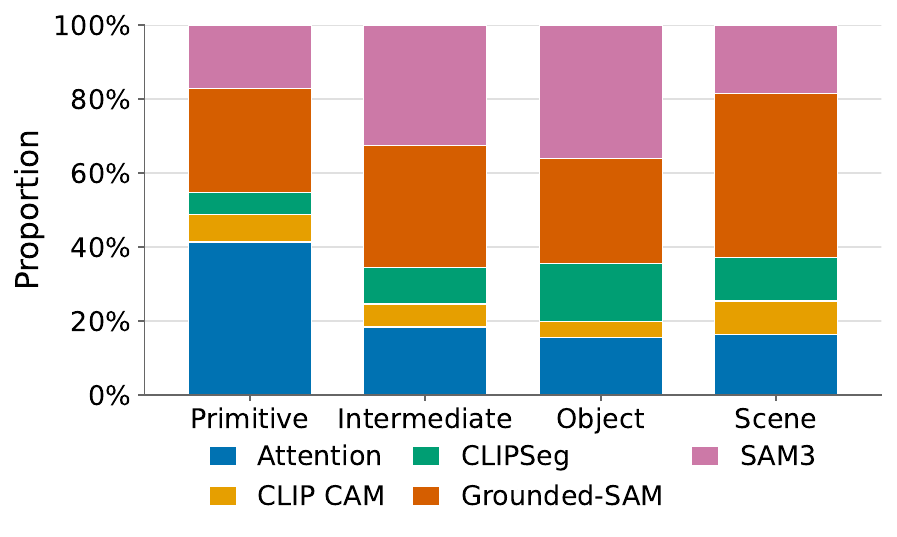}
        \vspace{-20pt}
        \caption{
Various distribution of selected annotation sources across concept levels. No single annotation tool is sufficient for all concepts, motivating the use of a heterogeneous proposal toolset.
}
        \label{fig:annotation_tool}
    \end{minipage}
    \label{fig:dataset_statistics_examples}
\end{figure}
\noindent \textbf{Verifier.}
Because each concept can receive multiple candidate masks with different levels of coverage, precision, and semantic alignment, we use a panel of multimodal verifiers to select reliable annotations. Specifically, we use three MLLMs, namely Qwen3.5-9B~\cite{qwen35blog}, InternVL3.5-8B~\cite{wang2025internvl35}, and GLM-4.6V-Flash~\cite{vteam2025glm45vglm41vthinkingversatilemultimodal}, as independent verifiers. Given the original image, the target concept, and overlay visualizations of the candidate masks, each verifier either selects the best candidate or rejects all candidates if none is acceptable. The decision is based on whether the candidate correctly grounds the concept, is spatially precise and sufficiently covers the relevant visual evidence. We use majority voting to obtain the final decision: a generated annotation is kept only if it is selected by at least two verifiers. This conservative voting scheme reduces model-specific verifier noise and prevents weak or semantically mismatched masks from being forced into the dataset. The final annotation manifest is obtained by combining all accepted generated masks with the direct source masks, yielding pixel-level rationales over the completed VIGOR concept set. See more details on verifier in \S Appendix~\ref{app:verifier}.

\textit{Do the choices of verifier models create circularity?}
Although some verifier families overlap with models evaluated in the benchmark, this does not create direct circularity in the benchmark. First, verifier judgments operate only at the output level: verifiers select or reject independently proposed candidate mask overlays, rather than generating the model rationales used in the benchmark. Second, majority voting across three verifiers reduces bias toward any single model family.

Figure~\ref{fig:annotation_tool} shows that selected proposal sources differ substantially across concept levels, indicating that different concepts require different grounding mechanisms. Primitive concepts rely more on attention-based proposals, which are better suited for diffuse cues such as colors, edges, and curves. Intermediate and object concepts are more often selected from segmentation-based methods such as Grounded-SAM and SAM3, which provide sharper masks for textures, materials, objects, and parts. Scene concepts are dominated by Grounded-SAM, suggesting that actions and spatial relations are often grounded through text-conditioned localization of entities and interaction regions. Overall, the distribution supports using a heterogeneous annotator pool, since no single proposal source is uniformly optimal across the concept hierarchy.

\begin{wrapfigure}{r}{0.5\textwidth}
\vspace{-10pt}
\centering
\includegraphics[width=\linewidth]{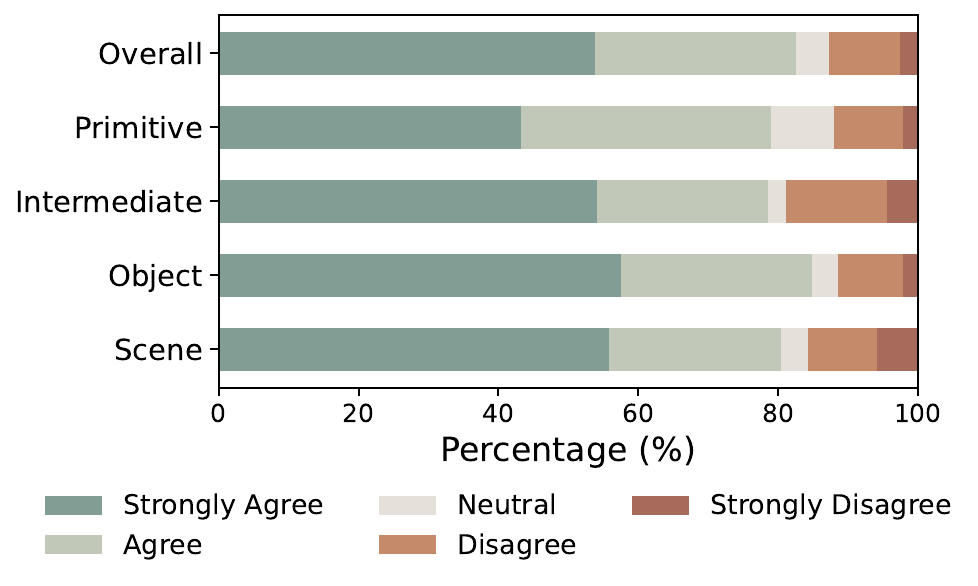}
\caption{
Human evaluation of VIGOR annotation quality across concept levels. Most annotations are rated as \emph{Agree} or \emph{Strongly Agree}, indicating consistently high reliability of the propose-then-verify pipeline across different concept levels.
}
\vspace{-10pt}
\label{fig:human_eval}
\end{wrapfigure}
\noindent \textbf{Human evaluation.}
To assess the automatic verification pipeline, we conduct a human evaluation on 500 randomly sampled verification results. Figure~\ref{fig:human_eval} summarizes ratings from three evaluators using a five-point Likert scale. Overall, the annotations receive strong human validation: $53.87\%$ of ratings are \emph{strongly agree} and $28.73\%$ are \emph{agree}, yielding $82.60\%$ positive ratings in total. Only $4.73\%$ of ratings are neutral, while negative judgments account for $12.67\%$. Among human--verifier disagreements, $55.8\%$ are verifier false negatives (valid candidates rejected) and $44.2\%$ false positives, confirming that the panel errs toward rejection, as intended by its conservative design. Across concept levels, object-level annotations obtain the highest positive agreement rate ($84.88\%$), consistent with objects being visually well-defined. Primitive, intermediate, and scene-level concepts remain more challenging, with positive agreement rates of $78.98\%$, $78.62\%$, and $80.39\%$, respectively, reflecting greater ambiguity in low-level attributes, textures/materials, and relational or scene-level evidence. These results suggest that the annotation pipeline produces reliable visual rationales while exposing the inherent difficulty of grounding more abstract or diffuse concepts. See full human evaluation result and inter-reviewer agreement analysis in \S Appendix~\ref{app:human_eval}.

\subsection{Dataset Statistics}

Table~\ref{tab:dataset_statistics} summarizes the statistics of VIGOR. The dataset contains 1,000 images and 3,011 unique visual concepts, with an average of 31.4 masks per image. In total, VIGOR provides 31,433 pixel-level masks, combining 18,038 masks sourced from existing annotations and 13,395 masks generated by our annotation framework. This design preserves reliable supervision from existing datasets while extending annotation coverage to concepts that lack direct pixel-level masks.

The masks span multiple levels of visual abstraction. Object-level concepts form the largest portion, with 23,863 masks, reflecting the central role of objects in visual understanding and MLLM outputs. Scene-level concepts contribute 4,363 masks, covering higher-level evidence such as relations and scene context. Primitive-level concepts account for 1,993 masks, capturing low-level visual properties such as color and edges, while intermediate-level concepts account for 1,214 masks, covering textures, materials, and shapes. This distribution shows that VIGOR is not limited to object-centric grounding, but provides pixel-level rationale supervision across a broader hierarchy of visual concepts.
\section{Experiments}

\begin{table}
\caption{
Main results on VIGOR benchmark. We report AP and Pointing Accuracy for overall grounding and for primitive, intermediate, object, and scene-level concepts. The best performing model in each column is \textbf{bolded} and the second best in \underline{underlined}.
}
\label{tab:main_results}
\centering
\resizebox{0.9\textwidth}{!}{
\begin{tabular}{lcccccccccc}
\toprule
\multirow{2}{*}{Model} & \multicolumn{2}{c}{Overall} & \multicolumn{2}{c}{Primitive} & \multicolumn{2}{c}{Intermediate} & \multicolumn{2}{c}{Object} & \multicolumn{2}{c}{Scene} \\
\cmidrule(lr){2-3} \cmidrule(lr){4-5} \cmidrule(lr){6-7} \cmidrule(lr){8-9} \cmidrule(lr){10-11}
 & AP & Pointing & AP & Pointing & mAP & Pointing & mAP & Pointing & mAP & Pointing \\
\midrule
Random & 5.8 & 5.7 & 4.0 & 3.9 & 1.9 & 1.8 & 5.0 & 4.9 & 17.7 & 17.3 \\
\midrule
InternVL3-1B & 19.6 & 27.7 & 28.7 & 40.7 & 9.9 & 14.1 & 22.3 & 32.9 & 24.9 & 30.5 \\
InternVL3.5-1B & 15.0 & 13.7 & 14.5 & 12.3 & 8.0 & 9.0 & 17.0 & 15.6 & 22.4 & 18.1 \\
\midrule
InternVL3-2B & 20.1 & 27.7 & 26.6 & 39.0 & 10.9 & 15.3 & 23.1 & 32.8 & 25.0 & 28.9 \\
InternVL3.5-2B & 28.8 & 36.5 & 36.9 & 46.9 & 16.8 & 21.3 & 34.0 & 43.7 & 31.0 & 37.2 \\
Qwen2-VL-2B & 19.4 & 23.8 & 26.3 & 37.0 & 12.0 & 15.3 & 21.8 & 26.8 & 22.9 & 23.8 \\
Qwen3-VL-2B & 25.2 & 31.9 & 30.9 & 41.5 & 15.7 & 20.5 & 29.4 & 37.2 & 27.5 & 32.2 \\
\midrule
Ministral3-3B & 21.1 & 31.5 & 28.2 & 41.0 & 10.2 & 16.2 & 24.1 & 37.5 & 28.5 & 36.9 \\
Qwen2.5-VL-3B & 27.5 & 34.6 & 36.1 & 45.6 & 17.2 & 21.5 & 31.9 & 40.8 & 28.6 & 34.7 \\
Gemma3-4B & 26.1 & 34.2 & 32.8 & 41.1 & 13.0 & 16.9 & 31.2 & 42.1 & 31.2 & 38.7 \\
InternVL3.5-4B & 29.7 & 37.2 & 37.6 & 45.9 & 16.9 & 20.9 & 35.3 & 44.7 & 32.3 & 39.2 \\
Qwen3-VL-4B & 28.3 & 34.5 & 34.9 & 43.5 & \underline{18.5} & \underline{23.2} & 33.1 & 40.7 & 29.0 & 32.2 \\
\midrule
Qwen2-VL-7B & 22.0 & 27.1 & 29.8 & 38.7 & 13.5 & 17.6 & 25.0 & 30.6 & 25.0 & 28.1 \\
Qwen2.5-VL-7B & 28.4 & 35.6 & 37.3 & 47.0 & 17.4 & 22.0 & 33.3 & 41.6 & 29.9 & 37.3 \\
InternVL3-8B & 26.6 & 35.2 & 34.0 & 45.1 & 15.4 & 20.0 & 31.5 & 42.7 & 28.7 & 34.7 \\
InternVL3.5-8B & \underline{30.6} & \underline{38.2} & \textbf{39.6} & \underline{48.2} & 17.3 & 21.7 & \underline{36.2} & \underline{45.6} & \underline{33.3} & \textbf{40.8} \\
Ministral3-8B & 22.9 & 32.0 & 30.6 & 40.7 & 10.9 & 16.3 & 26.4 & 38.4 & 30.0 & 37.3 \\
Qwen3-VL-8B & 26.3 & 32.2 & 31.1 & 40.4 & 17.7 & 22.1 & 30.5 & 37.5 & 27.4 & 31.0 \\
\midrule
Gemma3-12B & 27.8 & 35.2 & 35.8 & 43.3 & 15.2 & 19.2 & 32.8 & 42.4 & 31.7 & 38.3 \\
InternVL3-14B & 28.7 & 36.7 & 36.5 & 46.5 & 16.5 & 21.2 & 34.0 & 44.0 & 31.6 & 37.9 \\
InternVL3.5-14B & \textbf{31.2} & 38.0 & \underline{39.2} & 47.0 & 17.7 & 21.6 & \textbf{36.9} & 45.5 & \textbf{34.2} & \underline{39.9} \\
Ministral3-14B & 23.0 & 31.8 & 30.0 & 39.0 & 11.4 & 16.3 & 26.8 & 38.0 & 29.6 & 37.7 \\
\midrule
Gemma3-27B & 26.6 & 32.3 & 34.7 & 41.2 & 14.2 & 17.5 & 31.1 & 38.6 & 31.5 & 35.1 \\
Qwen2.5-VL-32B & 27.1 & 35.8 & 35.0 & 44.8 & 16.2 & 21.4 & 31.6 & 42.5 & 29.6 & 38.0 \\
Qwen3-VL-32B & 28.8 & 35.6 & 33.8 & 41.7 & \textbf{19.3} & \textbf{23.9} & 33.7 & 42.3 & 29.2 & 33.6 \\
InternVL3-38B & 27.0 & 35.7 & 33.9 & 45.0 & 16.1 & 20.7 & 31.8 & 43.1 & 29.5 & 35.8 \\
InternVL3.5-38B & 30.0 & \textbf{38.7} & 37.6 & \textbf{48.9} & 17.9 & 22.8 & 35.7 & \textbf{46.6} & 31.4 & 38.3 \\
\bottomrule
\end{tabular}
}
\end{table}
\subsection{Experimental Setup}
We evaluate representative open-source MLLMs from the InternVL~\cite{zhu2025internvl3,wang2025internvl35}, Qwen-VL~\cite{wang2024qwen2,bai2025qwen25vl,bai2025qwen3}, Gemma~\cite{gemma2025gemma3}, and Ministral~\cite{liu2026ministral} families, spanning model scales from 1B to 38B. For each concept, we extract its rationale map over visual tokens and compare it with the corresponding ground-truth mask. We also include a random baseline, whose rationale maps are sampled uniformly from $[0,1]$. We report two metrics: \textit{Average Precision}~\cite{choe2020evaluating} (AP), which measures rationale--mask alignment over varying thresholds, and \textit{Pointing Accuracy}~\cite{zhang2018top}, which measures whether the most-attended pixel falls inside the ground-truth mask. 
All models are evaluated without additional fine-tuning. Further experimental details are provided in Appendix~\ref{app:detailed_setup}.

Using model-generated captions would confound rationale correctness with caption coverage, since different models may mention different subsets of annotated concepts. We therefore construct a fixed caption for each VIGOR image from its annotated concept set, and evaluate all models on the same image--caption pair. This ensures that score differences reflect visual grounding quality rather than differences in generated content. The primary protocol thus measures \emph{caption-conditioned} rationale alignment; see \S Appendix~\ref{app:generated_caption} for additional evaluation results on model-generated captions. A human study confirms that the fixed captions are faithful and fluent ($4.86$ and $4.82$ on a five-point scale), and caption likelihood does not drive the results: a model's perplexity on the captions is only weakly and non-significantly correlated with its AP ($r=0.328$, $p=0.102$); see \S Appendix~\ref{app:caption_validation}. 
We compare several rationale extraction methods, including attention, GradCAM~\cite{selvaraju2020grad}, and LogitLens~\cite{jiang2024interpreting}, and use concept-token attention as our primary rationale proxy because it shows the strongest alignment with ground-truth rationale masks. See \S Appendix~\ref{app:additional_results} for benchmark results on other rationale extraction methods.

\subsection{Main Results}
\label{sec:main_results}


\noindent \textbf{Current MLLMs still have limited visual rationale correctness.}
Table~\ref{tab:main_results} shows that existing MLLMs remain weak at grounding the visual concepts they express. Even the best model obtains only 31.2 overall AP, while the highest overall Pointing score is 38.7. These results indicate that MLLMs can sometimes identify relevant visual regions, but their concept-level attention is far from reliably aligned with the annotated rationales. In particular, the relatively low Pointing scores suggest that even the most attended image region often falls outside the correct visual evidence. In other words, strong visual-language capability does not necessarily imply correct visual rationales: current MLLMs still only partially ``look at what they say.''

\noindent \textbf{MLLMs are better at coarse localization than full rationale alignment.}
Across almost all models and concept levels, Pointing is substantially higher than AP. This gap shows that models are often able to place their strongest attention peak near a relevant region, but the full attention distribution is not well aligned with the complete ground-truth rationale. For example, InternVL3.5-38B obtains 38.7 overall Pointing but only 30.0 overall AP; on object concepts, it reaches 46.6 Pointing but 35.7 AP. This pattern suggests that current MLLM rationales are spatially incomplete: they may identify a diagnostic part of the evidence while failing to cover the full region that supports the concept. Therefore, evaluating only peak localization would overestimate rationale correctness.

\begin{wrapfigure}{r}{0.44\textwidth}
\vspace{-18pt}
\centering
\includegraphics[width=\linewidth]{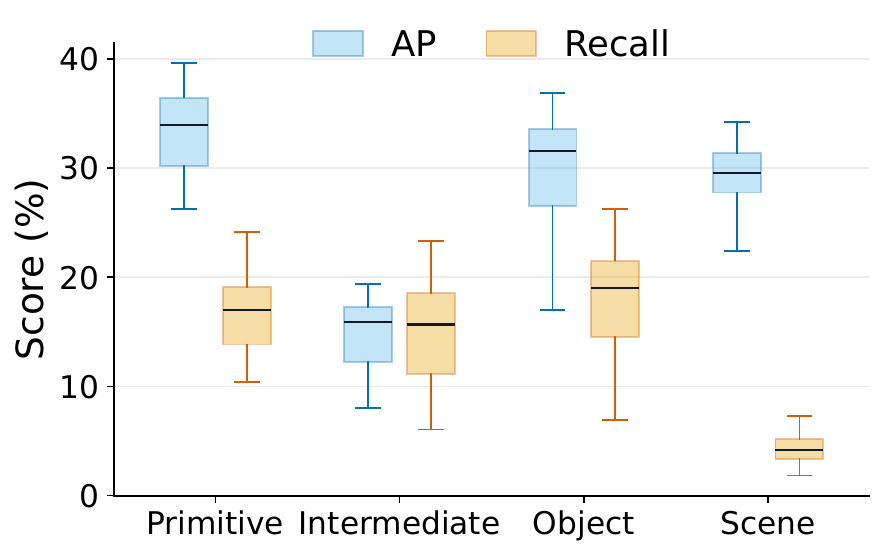}
\caption{
AP and Recall distributions across concept levels on VIGOR. The large recall gap, especially for scene concepts, shows that model rationales often cover only partial supporting evidence.
}
\vspace{-10pt}
\label{fig:concept_level}
\end{wrapfigure}
\noindent \textbf{Different concept levels expose different failure modes.}
As shown in Figure~\ref{fig:concept_level}, rationale quality varies substantially across concept levels. Recall is computed by normalizing each heatmap to $[0,1]$, binarizing it at $0.5$, and measuring ground-truth mask coverage. Primitive and object concepts achieve the highest AP but much lower Recall, indicating that models locate salient evidence while covering only part of the full rationale. Intermediate concepts have the weakest AP, suggesting that textures, materials, and shapes are difficult to separate from surrounding context. Scene concepts show the strongest under-coverage: AP is moderate, but Recall is below $5\%$, indicating reliance on a few diagnostic cues rather than global scene evidence. Overall, VIGOR reveals distinct failures: partial localization for primitive and object concepts, weak ranking for intermediate concepts, and severe under-coverage for scene concepts.

\subsection{Layerwise analysis}
\label{sec:layerwise_analysis}

\begin{wrapfigure}{r}{0.4\textwidth}
\centering
\includegraphics[width=\linewidth]{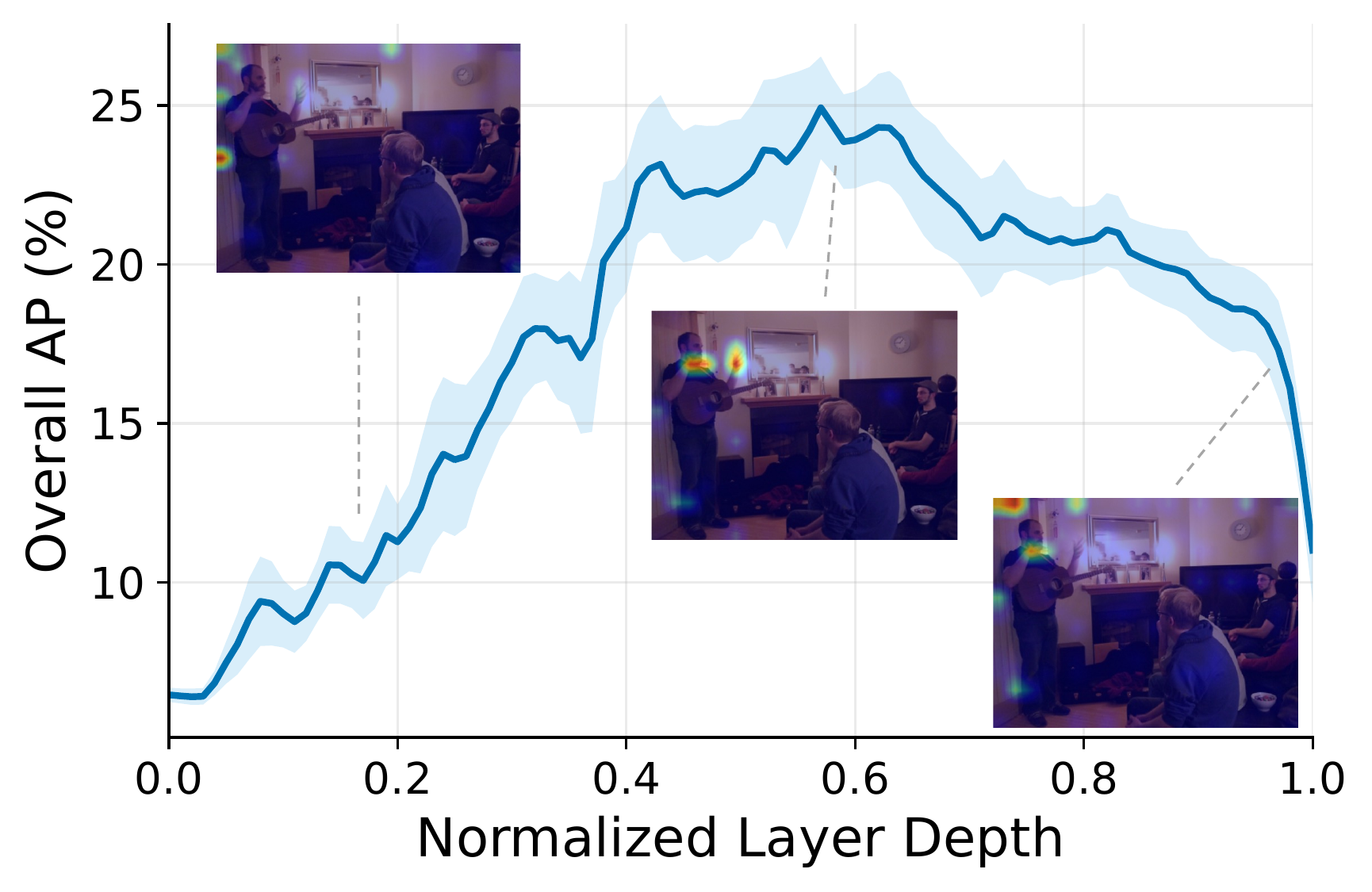}
\caption{
Layer-wise overall AP on VIGOR by normalized layer depth, averaged across all models.
}
\vspace{-10pt}
\label{fig:layerwise}
\end{wrapfigure}
Figure~\ref{fig:layerwise} shows the AP obtained from concept-token attention at different normalized layer depths. Rationale correctness follows a clear non-monotonic trend: AP is low in early layers, rises through the lower and middle layers, peaks around the middle-to-late depth, and then declines sharply near the final layers. This suggests that concept-grounded visual evidence is most spatially aligned after the model has integrated low-level visual features with higher-level semantic context, while final-layer attention is not necessarily the best proxy for visual rationale correctness. The qualitative examples show the same pattern: early attention maps are diffuse, middle-layer maps better localize the target evidence, and late-layer maps become sparse or incomplete. These results indicate that reliable visual rationales are concentrated in intermediate layers, and that evaluating only final-layer attention may miss important layer-wise grounding behavior.

\subsection{Rationale correctness correlates with hallucination robustness.}
\label{sec:hallucination}
\begin{wrapfigure}{r}{0.4\textwidth}
\vspace{-10pt}
\centering
\includegraphics[width=\linewidth]{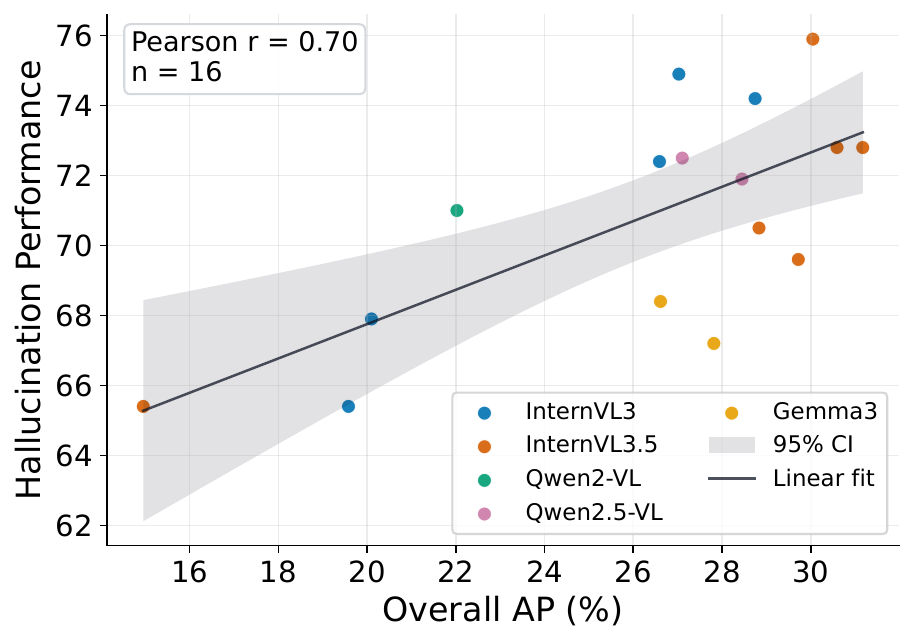}
\caption{
Relationship between rationale correctness and hallucination benchmark performance across 16 MLLMs. The positive correlation indicates that models with better visual rationale grounding also tend to achieve stronger hallucination robustness. Hallucination performances are sourced from~\cite{wang2025internvl35}.
}
\vspace{-10pt}
\label{fig:hallucination}
\end{wrapfigure}

To examine whether our benchmark captures a broader notion of trustworthiness, we compare each model's overall AP on VIGOR with its hallucination performance. Here, hallucination performance is computed as the average score across three hallucination benchmarks: HallusionBench~\cite{guan2024hallusionbench}, CRPE\cite{wang2024all}, and POPE~\cite{li2023pope}. As shown in Figure~\ref{fig:hallucination}, the two scores exhibit a clear positive correlation, with Pearson $r=0.70$ across 16 models. Models with lower rationale correctness generally obtain lower hallucination performance, while models with stronger visual grounding tend to be more robust against hallucination. This suggests that spatially faithful rationales are related to whether MLLMs produce visually grounded responses: models that better localize the concepts they process are also less likely to generate unsupported visual content.

The correlation is not perfect, indicating that hallucination robustness cannot be fully explained by attention-based grounding alone. Several models deviate from the overall trend, suggesting that language priors, instruction tuning and model family also affect hallucination behavior. Nevertheless, the positive relationship supports VIGOR as a complementary trustworthiness evaluation. Rather than only measuring whether models produce plausible answers, VIGOR evaluates whether the visual concepts processed by the model are grounded in appropriate image evidence.

\subsection{Grounding fine-tuning does not substantially improve rationale correctness.}
\label{sec:grounding_finetuning}
\begin{wraptable}{r}{0.4\textwidth}
\vspace{-15pt}
\centering
\caption{
Effect of grounding-oriented fine-tuning on VIGOR. Existing visual reasoning and grounding methods yield only marginal improvements in AP and Pointing Accuracy.
}
\label{tab:grounding_finetune}
\resizebox{0.95\linewidth}{!}{
\begin{tabular}{lcc}
\toprule
\textbf{Model} & \textbf{AP} & \textbf{Pointing} \\
\midrule
Qwen2.5-VL-3B & 27.5 & 34.6 \\
+VLM-R1    & 27.7 & 35.0 \\
\midrule
Qwen2.5-VL-7B & 28.4 & 35.6 \\
+Vision-R1 & 28.6 & 35.8 \\
\midrule
Qwen2.5-VL-7B & 28.4 & 35.6 \\
+ViGoRL    & 28.6 & 35.9 \\
\bottomrule
\end{tabular}
}
\vspace{-30pt}
\end{wraptable}

We further examine whether output-level grounding fine-tuning improves visual rationale correctness. As shown in Table~\ref{tab:grounding_finetune}, the gains are consistently marginal across AP and Pointing Accuracy. VLM-R1~\cite{shen2025vlm} improves Qwen2.5-VL-3B by only $0.2$ AP and $0.4$ Pointing Accuracy, while Vision-R1~\cite{zhan2025vision} and ViGoRL~\cite{sarch2025grounded} improve Qwen2.5-VL-7B by at most $0.2$ AP and $0.3$ Pointing Accuracy. These small changes indicate that existing grounding fine-tuning has little effect on the attention-based rationales evaluated by VIGOR.

This suggests that improving explicit grounding does not necessarily improve the model's visual rationale correctness. Grounding fine-tuning may mainly supervise output-level behavior, such as localized answers or grounded descriptions, without directly constraining whether concept-token attention aligns with the corresponding visual evidence. Moreover, many grounding datasets emphasize object-level localization, whereas VIGOR evaluates a broader concept hierarchy. These results indicate that better grounding outputs are not sufficient to guarantee rationale correctness.

\subsection{Error Analysis}
\label{sec:error_analysis}
\begin{figure}
  \centering
  \includegraphics[width=\linewidth]{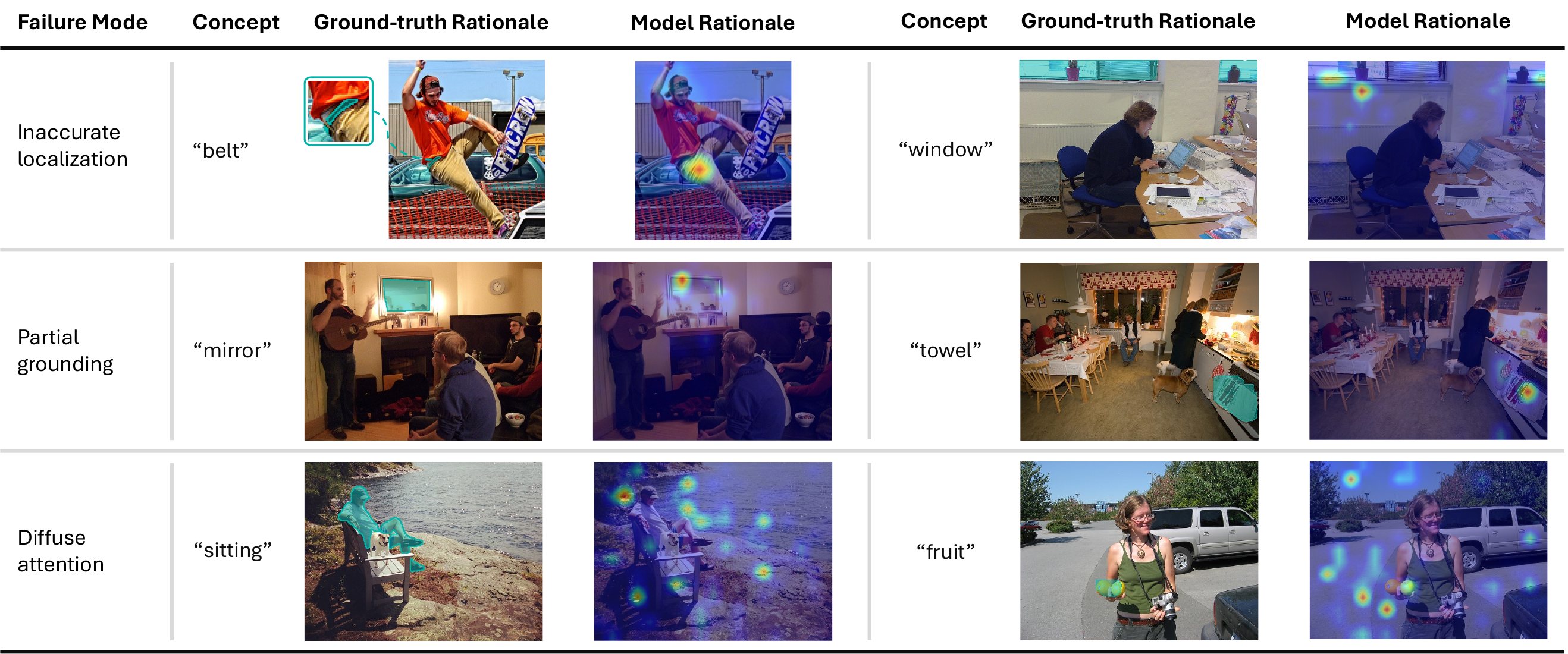}
  \caption{
Qualitative examples of rationale failures on VIGOR. Models can attend to incorrect regions, cover only part of the ground-truth evidence, or produce diffuse attention maps, showing that generated concepts are often not fully grounded in the appropriate visual rationale.
}
  \label{fig:error_analysis}
\end{figure}


Figure~\ref{fig:error_analysis} illustrates three common failure modes observed in MLLM visual rationales: inaccurate localization, partial grounding, and diffuse attention. These errors show that MLLMs often produce visual evidence that is semantically related to the target concept but spatially misaligned with the actual supporting region. Such failures are not always captured by answer-level evaluation, since the model may still generate a correct description while relying on incomplete or incorrect visual evidence. The examples suggest that current MLLMs often capture coarse semantic associations, but struggle to provide rationales that are spatially precise, complete, and selective.

\noindent \textbf{Inaccurate localization.}
The first failure mode occurs when the model attends to a visually related but spatially incorrect region. This is especially common for fine-grained concepts such as parts, attributes, or small objects, where the target occupies only a limited image area. In these cases, the model may identify the correct semantic neighborhood, such as the relevant person or object, but fail to localize the exact supporting evidence. This suggests that MLLM rationales often preserve coarse semantic association while lacking the spatial precision required for faithful concept grounding.

\noindent \textbf{Partial grounding.}
The second failure mode occurs when the model attends to only a subset of the target concept or its surrounding context. Unlike completely incorrect localization, partial grounding indicates that the model has identified the relevant region but fails to cover the full visual evidence. This can lead to high relevance but low completeness, particularly for objects or regions with extended spatial support. Such errors align with our quantitative results, where recall-oriented metrics remain low: the rationale may overlap with the target, but still omit substantial concept-relevant evidence.

\noindent \textbf{Diffuse attention.}
The third failure mode occurs when the model distributes attention broadly across the image rather than concentrating on the regions that support the target concept. This is particularly problematic for action, motion, and relation concepts, whose evidence depends on multiple entities and their spatial configuration. Instead of capturing this structured evidence, the model may spread attention over salient objects or background regions. Diffuse attention therefore reflects a failure to distinguish concept-relevant evidence from general visual saliency, limiting the reliability of attention-based rationales for compositional concepts.
\section{Conclusion}

VIGOR introduces a benchmark for evaluating whether MLLMs visually ground the concepts they express. Unlike standard evaluations that focus primarily on final-answer correctness, VIGOR pairs each image with concept-level pixel masks that specify the visual evidence expected to support descriptions, explanations, and other generated outputs. To construct these annotations at scale, we develop an automatic annotation framework that integrates dense source annotations, vocabulary-guided concept completion, candidate mask proposal, and conservative verifier voting. Our initial benchmark results show that current MLLMs and attribution methods often fail to localize the visual concepts they mention, with especially large gaps for relations, motion-like predicates, and other higher-level visual concepts. These findings suggest that output correctness alone provides an incomplete view of MLLM reliability, and that visually grounded rationale evaluation is necessary for measuring whether models are correct for the right visual reasons.

\section*{Acknowledgements}
This work is supported by the National Science Foundation under grant numbers CAREER 2340074, SLES 2416937, III CORE 2412675 and National Institutes of Health under grant number R21CA301093. Any opinions, findings and conclusions or recommendations expressed in this material are those of the authors and do not reflect the views of the supporting entities.

\newpage
\bibliographystyle{plain}
\bibliography{main}

@misc{bai2025qwen25vl,
  title = {Qwen2.5-{VL} Technical Report},
  author = {Bai, Shuai and Chen, Keqin and Liu, Xuejing and Wang, Jialin and Ge, Wenbin and Song, Sibo and Dang, Kai and Wang, Peng and Wang, Shijie and Tang, Jun and Zhong, Humen and Zhu, Yuanzhi and Yang, Mingkun and Li, Zhaohai and Wan, Jianqiang and Wang, Pengfei and Ding, Wei and Fu, Zheren and Xu, Yiheng and Ye, Jiabo and Zhang, Xi and Xie, Tianbao and Cheng, Zesen and Zhang, Hang and Yang, Zhibo and Xu, Haiyang and Lin, Junyang},
  year = {2025},
  eprint = {2502.13923},
  archivePrefix = {arXiv},
  primaryClass = {cs.CV}
}

@inproceedings{chefer2021transformer,
  title = {Transformer Interpretability Beyond Attention Visualization},
  author = {Chefer, Hila and Gur, Shir and Wolf, Lior},
  booktitle = {Proceedings of the IEEE/CVF Conference on Computer Vision and Pattern Recognition},
  pages = {782--791},
  year = {2021}
}

@misc{gemma2025gemma3,
  title = {Gemma 3 Technical Report},
  author = {{Gemma Team} and Kamath, Aishwarya and Ferret, Johan and Pathak, Shreya and Vieillard, Nino and Merhej, Ramona and Perrin, Sarah and Matejovicova, Tatiana and Rame, Alexandre and Riviere, Morgane and others},
  year = {2025},
  eprint = {2503.19786},
  archivePrefix = {arXiv},
  primaryClass = {cs.CL}
}

@inproceedings{gupta2019lvis,
  title = {{LVIS}: A Dataset for Large Vocabulary Instance Segmentation},
  author = {Gupta, Agrim and Dollar, Piotr and Girshick, Ross},
  booktitle = {Proceedings of the IEEE/CVF Conference on Computer Vision and Pattern Recognition},
  pages = {5356--5364},
  year = {2019}
}

@article{krishna2017visual,
  title = {Visual Genome: Connecting Language and Vision Using Crowdsourced Dense Image Annotations},
  author = {Krishna, Ranjay and Zhu, Yuke and Groth, Oliver and Johnson, Justin and Hata, Kenji and Kravitz, Joshua and Chen, Stephanie and Kalantidis, Yannis and Li, Li-Jia and Shamma, David A. and Bernstein, Michael S. and Fei-Fei, Li},
  journal = {International Journal of Computer Vision},
  volume = {123},
  number = {1},
  pages = {32--73},
  year = {2017},
  doi = {10.1007/s11263-016-0981-7}
}

@inproceedings{lin2014microsoft,
  title = {Microsoft {COCO}: Common Objects in Context},
  author = {Lin, Tsung-Yi and Maire, Michael and Belongie, Serge and Hays, James and Perona, Pietro and Ramanan, Deva and Dollar, Piotr and Zitnick, C. Lawrence},
  booktitle = {European Conference on Computer Vision},
  pages = {740--755},
  year = {2014}
}

@inproceedings{liu2023visual,
  title = {Visual Instruction Tuning},
  author = {Liu, Haotian and Li, Chunyuan and Wu, Qingyang and Lee, Yong Jae},
  booktitle = {Advances in Neural Information Processing Systems},
  volume = {36},
  pages = {34892--34916},
  year = {2023}
}

@inproceedings{luddecke2022image,
  title = {Image Segmentation Using Text and Image Prompts},
  author = {Luddecke, Timo and Ecker, Alexander},
  booktitle = {Proceedings of the IEEE/CVF Conference on Computer Vision and Pattern Recognition},
  pages = {7086--7096},
  year = {2022}
}

@inproceedings{ramanathan2023paco,
  title = {{PACO}: Parts and Attributes of Common Objects},
  author = {Ramanathan, Vignesh and Kalia, Anmol and Petrovic, Vladan and Wen, Yi and Zheng, Baixue and Guo, Baishan and Wang, Rui and Marquez, Aaron and Kovvuri, Rama and Kadian, Abhishek and Mousavi, Amir and Song, Yiwen and Dubey, Abhimanyu and Mahajan, Dhruv},
  booktitle = {Proceedings of the IEEE/CVF Conference on Computer Vision and Pattern Recognition},
  pages = {7141--7151},
  year = {2023}
}

@article{wang2025internvl35,
  title = {{InternVL3.5}: Advancing Open-Source Multimodal Models in Versatility, Reasoning, and Efficiency},
  author = {Wang, Weiyun and Gao, Zhangwei and Gu, Lixin and Pu, Hengjun and Cui, Long and Wei, Xingguang and Liu, Zhaoyang and Jing, Linglin and Ye, Shenglong and Shao, Jie and others},
  journal = {arXiv preprint arXiv:2508.18265},
  year = {2025}
}

@inproceedings{lai2024lisa,
  title={Lisa: Reasoning segmentation via large language model},
  author={Lai, Xin and Tian, Zhuotao and Chen, Yukang and Li, Yanwei and Yuan, Yuhui and Liu, Shu and Jia, Jiaya},
  booktitle={Proceedings of the IEEE/CVF conference on computer vision and pattern recognition},
  pages={9579--9589},
  year={2024}
}

@inproceedings{zhang2024llava,
  title={Llava-grounding: Grounded visual chat with large multimodal models},
  author={Zhang, Hao and Li, Hongyang and Li, Feng and Ren, Tianhe and Zou, Xueyan and Liu, Shilong and Huang, Shijia and Gao, Jianfeng and Leizhang and Li, Chunyuan and others},
  booktitle={European Conference on Computer Vision},
  pages={19--35},
  year={2024},
  organization={Springer}
}

@inproceedings{ren2024pixellm,
  title={Pixellm: Pixel reasoning with large multimodal model},
  author={Ren, Zhongwei and Huang, Zhicheng and Wei, Yunchao and Zhao, Yao and Fu, Dongmei and Feng, Jiashi and Jin, Xiaojie},
  booktitle={Proceedings of the IEEE/CVF Conference on Computer Vision and Pattern Recognition},
  pages={26374--26383},
  year={2024}
}

@inproceedings{ribeiro2016should,
  title={``Why should i trust you?'' Explaining the predictions of any classifier},
  author={Ribeiro, Marco Tulio and Singh, Sameer and Guestrin, Carlos},
  booktitle={Proceedings of the 22nd ACM SIGKDD international conference on knowledge discovery and data mining},
  pages={1135--1144},
  year={2016}
}

@article{geirhos2020shortcut,
  title={Shortcut Learning in Deep Neural Networks},
  author={Geirhos, Robert and Jacobsen, J{\"o}rn-Henrik and Michaelis, Claudio and Zemel, Richard and Brendel, Wieland and Bethge, Matthias and Wichmann, Felix A.},
  journal={Nature Machine Intelligence},
  volume={2},
  pages={665--673},
  year={2020}
}

@inproceedings{goyal2017making,
  title={Making the V in VQA Matter: Elevating the Role of Image Understanding in Visual Question Answering},
  author={Goyal, Yash and Khot, Tejas and Summers-Stay, Douglas and Batra, Dhruv and Parikh, Devi},
  booktitle={Proceedings of the IEEE Conference on Computer Vision and Pattern Recognition},
  year={2017}
}

@inproceedings{li2023blip2,
  title={BLIP-2: Bootstrapping Language-Image Pre-training with Frozen Image Encoders and Large Language Models},
  author={Li, Junnan and Li, Dongxu and Savarese, Silvio and Hoi, Steven},
  booktitle={International Conference on Machine Learning},
  year={2023}
}

@inproceedings{zhu2023minigpt4,
  title={MiniGPT-4: Enhancing Vision-Language Understanding with Advanced Large Language Models},
  author={Zhu, Deyao and Chen, Jun and Shen, Xiaoqian and Li, Xiang and Elhoseiny, Mohamed},
  booktitle={International Conference on Learning Representations},
  year={2024}
}

@inproceedings{mahmood2025phrase,
  title={Phrase-grounded Fact-checking for Automatically Generated Chest X-ray Reports},
  author={Mahmood, Razi and Machado-Reyes, Diego and Wu, Joy and Kaviani, Parisa and Wong, Ken C. L. and D'Souza, Niharika and Kalra, Mannudeep and Wang, Ge and Yan, Pingkun and Syeda-Mahmood, Tanveer},
  booktitle={Proceedings of Medical Image Computing and Computer Assisted Intervention},
  pages={441--452},
  year={2025},
  publisher={Springer Nature Switzerland}
}

@article{degrave2021ai,
  title={AI for Radiographic COVID-19 Detection Selects Shortcuts over Signal},
  author={DeGrave, Alex J. and Janizek, Joseph D. and Lee, Su-In},
  journal={Nature Machine Intelligence},
  volume={3},
  pages={610--619},
  year={2021},
  doi={10.1038/s42256-021-00338-7}
}

@inproceedings{
fu2026mme,
title={{MME}: A Comprehensive Evaluation Benchmark for Multimodal Large Language Models},
author={Chaoyou Fu and Peixian Chen and Yunhang Shen and Yulei Qin and Mengdan Zhang and Xu Lin and Jinrui Yang and Xiawu Zheng and Ke Li and Xing Sun and Yunsheng Wu and Rongrong Ji and Caifeng Shan and Ran He},
booktitle={The Thirty-ninth Annual Conference on Neural Information Processing Systems Datasets and Benchmarks Track},
year={2026},
}

@inproceedings{liu2024mmbench,
  title={MMBench: Is Your Multi-modal Model an All-around Player?},
  author={Liu, Yuan and Duan, Haodong and Zhang, Yuanhan and Li, Bo and Zhang, Songyang and Zhao, Wangbo and Yuan, Yike and Wang, Jiaqi and He, Conghui and Liu, Ziwei and Chen, Kai and Lin, Dahua},
  booktitle={European Conference on Computer Vision},
  year={2024}
}

@inproceedings{li2024seed,
  title={SEED-Bench: Benchmarking Multimodal Large Language Models},
  author={Li, Bohao and Wang, Rui and Wang, Guangzhi and Ge, Yuying and Ge, Yixiao and Shan, Ying},
  booktitle={Proceedings of the IEEE/CVF Conference on Computer Vision and Pattern Recognition},
  year={2024}
}

@inproceedings{yue2024mmmu,
  title={MMMU: A Massive Multi-discipline Multimodal Understanding and Reasoning Benchmark for Expert AGI},
  author={Yue, Xiang and Ni, Yuansheng and Zhang, Kai and Zheng, Tianyu and Liu, Ruoqi and Zhang, Ge and Stevens, Samuel and Jiang, Dongfu and Ren, Weiming and Sun, Yuxuan and Wei, Cong and Yu, Botao and Yuan, Ruibin and Sun, Renliang and Yin, Ming and Zheng, Boyuan and Yang, Zhenzhu and Liu, Yibo and Huang, Wenhao and Sun, Huan and Su, Yu},
  booktitle={Proceedings of the IEEE/CVF Conference on Computer Vision and Pattern Recognition},
  year={2024}
}

@inproceedings{li2023pope,
  title={Evaluating Object Hallucination in Large Vision-Language Models},
  author={Li, Yifan and Du, Yifan and Zhou, Kun and Wang, Jinpeng and Zhao, Wayne Xin and Wen, Ji-Rong},
  booktitle={Proceedings of the 2023 Conference on Empirical Methods in Natural Language Processing},
  year={2023}
}

@inproceedings{guan2024hallusionbench,
  title={HallusionBench: An Advanced Diagnostic Suite for Entangled Language Hallucination and Visual Illusion in Large Vision-Language Models},
  author={Guan, Tianrui and Liu, Fuxiao and Wu, Xiyang and Xian, Ruiqi and Li, Zongxia and Liu, Xiaoyu and Wang, Xijun and Chen, Lichang and Huang, Furong and Yacoob, Yaser and Manocha, Dinesh and Zhou, Tianyi},
  booktitle={Proceedings of the IEEE/CVF Conference on Computer Vision and Pattern Recognition},
  year={2024}
}

@inproceedings{yu2023mmvet,
  title={MM-Vet: Evaluating Large Multimodal Models for Integrated Capabilities},
  author={Yu, Weihao and Yang, Zhengyuan and Li, Linjie and Wang, Jianfeng and Lin, Kevin and Liu, Zicheng and Wang, Xinchao and Wang, Lijuan},
  booktitle={arXiv preprint arXiv:2308.02490},
  year={2023}
}

@inproceedings{yu2016modeling,
  title={Modeling Context in Referring Expressions},
  author={Yu, Licheng and Poirson, Patrick and Yang, Shan and Berg, Alexander C. and Berg, Tamara L.},
  booktitle={European Conference on Computer Vision},
  pages={69--85},
  year={2016}
}

@inproceedings{kazemzadeh2014referitgame,
  title={ReferItGame: Referring to Objects in Photographs of Natural Scenes},
  author={Kazemzadeh, Sahar and Ordonez, Vicente and Matten, Mark and Berg, Tamara L.},
  booktitle={Proceedings of the 2014 Conference on Empirical Methods in Natural Language Processing},
  pages={787--798},
  year={2014}
}

@inproceedings{rasheed2024glamm,
  title={Glamm: Pixel grounding large multimodal model},
  author={Rasheed, Hanoona and Maaz, Muhammad and Shaji, Sahal and Shaker, Abdelrahman and Khan, Salman and Cholakkal, Hisham and Anwer, Rao M and Xing, Eric and Yang, Ming-Hsuan and Khan, Fahad S},
  booktitle={Proceedings of the IEEE/CVF Conference on Computer Vision and Pattern Recognition},
  pages={13009--13018},
  year={2024}
}

@inproceedings{wu2020phrasecut,
  title={Phrasecut: Language-based image segmentation in the wild},
  author={Wu, Chenyun and Lin, Zhe and Cohen, Scott and Bui, Trung and Maji, Subhransu},
  booktitle={Proceedings of the IEEE/CVF Conference on Computer Vision and Pattern Recognition},
  pages={10216--10225},
  year={2020}
}

@inproceedings{park2018multimodal,
  title={Multimodal explanations: Justifying decisions and pointing to the evidence},
  author={Park, Dong Huk and Hendricks, Lisa Anne and Akata, Zeynep and Rohrbach, Anna and Schiele, Bernt and Darrell, Trevor and Rohrbach, Marcus},
  booktitle={Proceedings of the IEEE conference on computer vision and pattern recognition},
  pages={8779--8788},
  year={2018}
}

@article{das2017human,
  title={Human attention in visual question answering: Do humans and deep networks look at the same regions?},
  author={Das, Abhishek and Agrawal, Harsh and Zitnick, Larry and Parikh, Devi and Batra, Dhruv},
  journal={Computer Vision and Image Understanding},
  volume={163},
  pages={90--100},
  year={2017},
  publisher={Elsevier}
}

@article{do2020snli,
  title={e-snli-ve: Corrected visual-textual entailment with natural language explanations},
  author={Do, Virginie and Camburu, Oana-Maria and Akata, Zeynep and Lukasiewicz, Thomas},
  journal={arXiv preprint arXiv:2004.03744},
  year={2020}
}

@inproceedings{rohrbach2018object,
  title={Object hallucination in image captioning},
  author={Rohrbach, Anna and Hendricks, Lisa Anne and Burns, Kaylee and Darrell, Trevor and Saenko, Kate},
  booktitle={Proceedings of the 2018 Conference on Empirical Methods in Natural Language Processing},
  pages={4035--4045},
  year={2018}
}

@inproceedings{caesar2018coco,
  title={Coco-stuff: Thing and stuff classes in context},
  author={Caesar, Holger and Uijlings, Jasper and Ferrari, Vittorio},
  booktitle={Proceedings of the IEEE conference on computer vision and pattern recognition},
  pages={1209--1218},
  year={2018}
}

@inproceedings{cimpoi2014describing,
  title={Describing textures in the wild},
  author={Cimpoi, Mircea and Maji, Subhransu and Kokkinos, Iasonas and Mohamed, Sammy and Vedaldi, Andrea},
  booktitle={Proceedings of the IEEE conference on computer vision and pattern recognition},
  pages={3606--3613},
  year={2014}
}

@inproceedings{zhou2017scene,
  title={Scene parsing through ade20k dataset},
  author={Zhou, Bolei and Zhao, Hang and Puig, Xavier and Fidler, Sanja and Barriuso, Adela and Torralba, Antonio},
  booktitle={Proceedings of the IEEE conference on computer vision and pattern recognition},
  pages={633--641},
  year={2017}
}

@inproceedings{pham2021learning,
  title={Learning to predict visual attributes in the wild},
  author={Pham, Khoi and Kafle, Kushal and Lin, Zhe and Ding, Zhihong and Cohen, Scott and Tran, Quan and Shrivastava, Abhinav},
  booktitle={Proceedings of the IEEE/CVF conference on computer vision and pattern recognition},
  pages={13018--13028},
  year={2021}
}

@article{kuznetsova2020open,
  title={The open images dataset v4: Unified image classification, object detection, and visual relationship detection at scale},
  author={Kuznetsova, Alina and Rom, Hassan and Alldrin, Neil and Uijlings, Jasper and Krasin, Ivan and Pont-Tuset, Jordi and Kamali, Shahab and Popov, Stefan and Malloci, Matteo and Kolesnikov, Alexander and others},
  journal={International journal of computer vision},
  volume={128},
  number={7},
  pages={1956--1981},
  year={2020},
  publisher={Springer}
}

@inproceedings{gu2018ava,
  title={Ava: A video dataset of spatio-temporally localized atomic visual actions},
  author={Gu, Chunhui and Sun, Chen and Ross, David A and Vondrick, Carl and Pantofaru, Caroline and Li, Yeqing and Vijayanarasimhan, Sudheendra and Toderici, George and Ricco, Susanna and Sukthankar, Rahul and others},
  booktitle={Proceedings of the IEEE conference on computer vision and pattern recognition},
  pages={6047--6056},
  year={2018}
}

@article{kay2017kinetics,
  title={The kinetics human action video dataset},
  author={Kay, Will and Carreira, Joao and Simonyan, Karen and Zhang, Brian and Hillier, Chloe and Vijayanarasimhan, Sudheendra and Viola, Fabio and Green, Tim and Back, Trevor and Natsev, Paul and others},
  journal={arXiv preprint arXiv:1705.06950},
  year={2017}
}

@inproceedings{hudson2019gqa,
  title={Gqa: A new dataset for real-world visual reasoning and compositional question answering},
  author={Hudson, Drew A and Manning, Christopher D},
  booktitle={Proceedings of the IEEE/CVF conference on computer vision and pattern recognition},
  pages={6700--6709},
  year={2019}
}

@article{carion2025sam,
  title={Sam 3: Segment anything with concepts},
  author={Carion, Nicolas and Gustafson, Laura and Hu, Yuan-Ting and Debnath, Shoubhik and Hu, Ronghang and Suris, Didac and Ryali, Chaitanya and Alwala, Kalyan Vasudev and Khedr, Haitham and Huang, Andrew and others},
  journal={arXiv preprint arXiv:2511.16719},
  year={2025}
}

@article{ren2024grounded,
  title={Grounded sam: Assembling open-world models for diverse visual tasks},
  author={Ren, Tianhe and Liu, Shilong and Zeng, Ailing and Lin, Jing and Li, Kunchang and Cao, He and Chen, Jiayu and Huang, Xinyu and Chen, Yukang and Yan, Feng and others},
  journal={arXiv preprint arXiv:2401.14159},
  year={2024}
}

@misc{qwen35blog,
    title = {Qwen3.5: Accelerating Productivity with Native Multimodal Agents},
    url = {https://qwen.ai/blog?id=qwen3.5},
    author = {Qwen Team},
    month = {February},
    year = {2026}
}

@misc{vteam2025glm45vglm41vthinkingversatilemultimodal,
      title={GLM-4.5V and GLM-4.1V-Thinking: Towards Versatile Multimodal Reasoning with Scalable Reinforcement Learning},
      author={V Team and Wenyi Hong and Wenmeng Yu and Xiaotao Gu and Guo Wang and Guobing Gan and Haomiao Tang and Jiale Cheng and Ji Qi and Junhui Ji and Lihang Pan and Shuaiqi Duan and Weihan Wang and Yan Wang and Yean Cheng and Zehai He and Zhe Su and Zhen Yang and Ziyang Pan and Aohan Zeng and Baoxu Wang and Bin Chen and Boyan Shi and Changyu Pang and Chenhui Zhang and Da Yin and Fan Yang and Guoqing Chen and Jiazheng Xu and Jiale Zhu and Jiali Chen and Jing Chen and Jinhao Chen and Jinghao Lin and Jinjiang Wang and Junjie Chen and Leqi Lei and Letian Gong and Leyi Pan and Mingdao Liu and Mingde Xu and Mingzhi Zhang and Qinkai Zheng and Sheng Yang and Shi Zhong and Shiyu Huang and Shuyuan Zhao and Siyan Xue and Shangqin Tu and Shengbiao Meng and Tianshu Zhang and Tianwei Luo and Tianxiang Hao and Tianyu Tong and Wenkai Li and Wei Jia and Xiao Liu and Xiaohan Zhang and Xin Lyu and Xinyue Fan and Xuancheng Huang and Yanling Wang and Yadong Xue and Yanfeng Wang and Yanzi Wang and Yifan An and Yifan Du and Yiming Shi and Yiheng Huang and Yilin Niu and Yuan Wang and Yuanchang Yue and Yuchen Li and Yutao Zhang and Yuting Wang and Yu Wang and Yuxuan Zhang and Zhao Xue and Zhenyu Hou and Zhengxiao Du and Zihan Wang and Peng Zhang and Debing Liu and Bin Xu and Juanzi Li and Minlie Huang and Yuxiao Dong and Jie Tang},
      year={2025},
      eprint={2507.01006},
      archivePrefix={arXiv},
      primaryClass={cs.CV},
      url={https://arxiv.org/abs/2507.01006}, 
}

@article{selvaraju2020grad,
  title={Grad-CAM: visual explanations from deep networks via gradient-based localization},
  author={Selvaraju, Ramprasaath R and Cogswell, Michael and Das, Abhishek and Vedantam, Ramakrishna and Parikh, Devi and Batra, Dhruv},
  journal={International journal of computer vision},
  volume={128},
  number={2},
  pages={336--359},
  year={2020},
  publisher={Springer}
}

@article{jiang2024interpreting,
  title={Interpreting and editing vision-language representations to mitigate hallucinations},
  author={Jiang, Nick and Kachinthaya, Anish and Petryk, Suzie and Gandelsman, Yossi},
  journal={arXiv preprint arXiv:2410.02762},
  year={2024}
}

@article{zhu2025internvl3,
  title={Internvl3: Exploring advanced training and test-time recipes for open-source multimodal models},
  author={Zhu, Jinguo and Wang, Weiyun and Chen, Zhe and Liu, Zhaoyang and Ye, Shenglong and Gu, Lixin and Tian, Hao and Duan, Yuchen and Su, Weijie and Shao, Jie and others},
  journal={arXiv preprint arXiv:2504.10479},
  year={2025}
}

@article{wang2024qwen2,
  title={Qwen2-vl: Enhancing vision-language model's perception of the world at any resolution},
  author={Wang, Peng and Bai, Shuai and Tan, Sinan and Wang, Shijie and Fan, Zhihao and Bai, Jinze and Chen, Keqin and Liu, Xuejing and Wang, Jialin and Ge, Wenbin and others},
  journal={arXiv preprint arXiv:2409.12191},
  year={2024}
}

@article{bai2025qwen3,
  title={Qwen3-vl technical report},
  author={Bai, Shuai and Cai, Yuxuan and Chen, Ruizhe and Chen, Keqin and Chen, Xionghui and Cheng, Zesen and Deng, Lianghao and Ding, Wei and Gao, Chang and Ge, Chunjiang and others},
  journal={arXiv preprint arXiv:2511.21631},
  year={2025}
}

@article{liu2026ministral,
  title={Ministral 3},
  author={Liu, Alexander H and Khandelwal, Kartik and Subramanian, Sandeep and Jouault, Victor and Rastogi, Abhinav and Sad{\'e}, Adrien and Jeffares, Alan and Jiang, Albert and Cahill, Alexandre and Gavaudan, Alexandre and others},
  journal={arXiv preprint arXiv:2601.08584},
  year={2026}
}

@inproceedings{choe2020evaluating,
  title={Evaluating weakly supervised object localization methods right},
  author={Choe, Junsuk and Oh, Seong Joon and Lee, Seungho and Chun, Sanghyuk and Akata, Zeynep and Shim, Hyunjung},
  booktitle={Proceedings of the IEEE/CVF conference on computer vision and pattern recognition},
  pages={3133--3142},
  year={2020}
}

@article{zhang2018top,
  title={Top-down neural attention by excitation backprop},
  author={Zhang, Jianming and Bargal, Sarah Adel and Lin, Zhe and Brandt, Jonathan and Shen, Xiaohui and Sclaroff, Stan},
  journal={International Journal of Computer Vision},
  volume={126},
  number={10},
  pages={1084--1102},
  year={2018},
  publisher={Springer}
}

@inproceedings{wang2024all,
  title={The all-seeing project v2: Towards general relation comprehension of the open world},
  author={Wang, Weiyun and Ren, Yiming and Luo, Haowen and Li, Tiantong and Yan, Chenxiang and Chen, Zhe and Wang, Wenhai and Li, Qingyun and Lu, Lewei and Zhu, Xizhou and others},
  booktitle={European Conference on Computer Vision},
  pages={471--490},
  year={2024},
  organization={Springer}
}

@article{shen2025vlm,
  title={Vlm-r1: A stable and generalizable r1-style large vision-language model},
  author={Shen, Haozhan and Liu, Peng and Li, Jingcheng and Fang, Chunxin and Ma, Yibo and Liao, Jiajia and Shen, Qiaoli and Zhang, Zilun and Zhao, Kangjia and Zhang, Qianqian and others},
  journal={arXiv preprint arXiv:2504.07615},
  year={2025}
}

@article{zhan2025vision,
  title={Vision-r1: Evolving human-free alignment in large vision-language models via vision-guided reinforcement learning},
  author={Zhan, Yufei and Zhu, Yousong and Zheng, Shurong and Zhao, Hongyin and Yang, Fan and Tang, Ming and Wang, Jinqiao},
  journal={arXiv preprint arXiv:2503.18013},
  year={2025}
}

@inproceedings{sarch2025grounded,
  title={Grounded Reinforcement Learning for Visual Reasoning},
  author={Sarch, Gabriel Herbert and Saha, Snigdha and Khandelwal, Naitik and Jain, Ayush and Tarr, Michael J. and Kumar, Aviral and Fragkiadaki, Katerina},
  booktitle={Advances in Neural Information Processing Systems},
  year={2025}
}

@inproceedings{vedantam2015cider,
  title={{CIDEr}: Consensus-based image description evaluation},
  author={Vedantam, Ramakrishna and Lawrence Zitnick, C and Parikh, Devi},
  booktitle={Proceedings of the IEEE Conference on Computer Vision and Pattern Recognition},
  pages={4566--4575},
  year={2015}
}

@inproceedings{lin2004rouge,
  title={{ROUGE}: A package for automatic evaluation of summaries},
  author={Lin, Chin-Yew},
  booktitle={Text Summarization Branches Out},
  pages={74--81},
  year={2004}
}

@article{chen2015microsoft,
  title={Microsoft {COCO} captions: Data collection and evaluation server},
  author={Chen, Xinlei and Fang, Hao and Lin, Tsung-Yi and Vedantam, Ramakrishna and Gupta, Saurabh and Doll{\'a}r, Piotr and Zitnick, C Lawrence},
  journal={arXiv preprint arXiv:1504.00325},
  year={2015}
}

\newpage
\appendix 
\section{Discussions}
\subsection{Limitations}
\label{app:limitations}
VIGOR provides pixel-level supervision for evaluating visually grounded rationales, but it has several limitations. First, our evaluation uses attention maps as a proxy for model rationales. Although attention provides a direct and model-accessible signal of how concept tokens interact with image tokens, it is not a complete causal explanation of model behavior. A model may attend to a region without using it for prediction, or rely on visual evidence that is not fully captured by attention. Therefore, our current experiments evaluate attention-based rationale alignment rather than definitive causal grounding, and our claims about model rationales should be read with this scope in mind. We stress that this limitation concerns the rationale-extraction method, not the benchmark itself: VIGOR's concept-level pixel annotations are model- and attribution-agnostic, and provide the ground-truth evidence against which attention, GradCAM, LogitLens, or future causal attribution methods can all be evaluated (\S Appendix~\ref{app:additional_results}).

Second, the annotation pipeline is automatic and may introduce noise. We combine existing dense annotations, proposal-based mask generation, and MLLM-based verification to scale beyond object-centric supervision. While verifier voting and conservative rejection reduce annotation errors, some masks may still be incomplete, over-extended, or semantically ambiguous, especially for abstract concepts such as relations, motion, and scene-level evidence. Human validation suggests high annotation quality, but it does not eliminate all possible errors.

Third, the concept vocabulary is broad but not exhaustive. VIGOR covers primitive, intermediate, object, and scene-level concepts, but natural images contain an open-ended space of visual properties and relations. Some fine-grained attributes, rare objects, domain-specific concepts, or culturally dependent semantics may be missing. As a result, the benchmark should be viewed as a representative evaluation of visual rationale grounding rather than a complete ontology of all visual evidence.

Fourth, VIGOR is limited in scale and image distribution. The benchmark contains 1,000 images selected from existing vision datasets, which enables dense concept-level annotation but may not reflect all real-world visual domains. Performance on VIGOR may not directly generalize to specialized settings such as medical imaging, remote sensing, embodied navigation, or low-quality user-generated images. We note that raw image count understates the effective scale of the benchmark: the 1,000 images are densely annotated with pixel-level evidence for 3,011 unique concepts across four abstraction levels, yielding more than 31K concept-level grounding evaluations, consistent with VIGOR's intended role as a diagnostic benchmark. Nevertheless, verifying that annotation quality holds when the pipeline is applied to substantially larger and more diverse image collections remains future work.

Fifth, our evaluation is conducted in the image-captioning setting. The primary protocol measures caption-conditioned rationale alignment on fixed captions, and Sec.~\ref{app:generated_caption} complements it with model-generated captions; neither covers other decision-making tasks such as visual question answering, where the concepts a model relies on may differ from those it mentions. The annotation pipeline and the resulting pixel-level rationales are not specific to captioning, and extending the evaluation protocol to such tasks is a natural next step, but we do not claim broad task coverage in this work.

Finally, our experiments evaluate a representative set of open-source MLLMs under a fixed prompting and attention-extraction protocol. Different prompts, decoding settings, model implementations, or attribution methods may lead to different grounding scores. Future work should extend VIGOR to causal attribution methods, larger and more diverse image collections, human-in-the-loop annotation, and task-conditioned evaluations where rationale correctness can be connected more directly to downstream model behavior.

\subsection{Broader Impact}
\label{app:broader_impact}
VIGOR aims to support more trustworthy evaluation of multimodal large language models by measuring whether their expressed visual concepts are grounded in appropriate image evidence. This can have positive societal impact by exposing cases where models produce plausible or correct outputs for the wrong visual reasons. Such failures are especially important in safety-critical or high-stakes domains, where a model that appears accurate may still rely on spurious cues, irrelevant regions, or dataset artifacts. By providing concept-level pixel annotations and rationale-grounding metrics, VIGOR can help researchers diagnose these failures and develop models whose visual reasoning is more transparent and reliable.

At the same time, VIGOR should not be interpreted as a complete safety certification. The benchmark evaluates grounding under a fixed protocol and primarily uses attention-based rationale maps, which are imperfect proxies for causal model behavior. High performance on VIGOR therefore does not guarantee that a model will always rely on correct evidence in deployment. There is a risk that benchmark scores could be overinterpreted as evidence of general trustworthiness, especially outside the image distributions, model families, and concept types covered by the dataset.

The release of VIGOR also raises responsible-use considerations. Because the benchmark is designed for evaluation rather than model generation, its direct misuse risk is limited. However, improved rationale-grounding evaluation could be used to optimize models for appearing visually grounded without necessarily improving their causal reasoning. To mitigate this risk, we document the benchmark assumptions, annotation pipeline, known limitations, and intended use cases. We encourage users to treat VIGOR as one component of a broader evaluation suite, complementing output correctness, hallucination, robustness, fairness, privacy, and causal attribution analyses.

VIGOR is constructed from existing vision datasets and derived annotations. We follow the licenses and terms of use of the source datasets and release only the assets permitted by those terms. The benchmark does not introduce new personal data collection or human-subject experimentation. Future extensions should further examine dataset bias, demographic representation, and domain coverage, particularly before using visually grounded rationale evaluation to support deployment decisions in sensitive applications.

\section{Details of the Automatic Annotation Framework}

\subsection{Concept Vocabulary Construction}
\label{app:concept_vocab}

Starting from the raw concept pool, we use \texttt{gpt-5.4}, \texttt{claude-sonnet-4-20250514}, and \texttt{gemini-2.5-pro} to filter candidate concepts. The filtering prompt is shown in Figures~\ref{prompt:prompt_filter_concept_1}--\ref{prompt:prompt_filter_concept_3}. To reduce long-context degradation, we process candidates in batches of 50. A concept is retained only if at least two of the three LLMs choose to keep it. After this filtering stage, we perform a final cleaning pass with \texttt{gpt-5.4} to remove residual duplicates, lexical variants, and inconsistent category assignments. The cleaning prompt is shown in Figures~\ref{prompt:prompt_clean_concept_1} and~\ref{prompt:prompt_clean_concept_2}. Unlike filtering, cleaning is performed over the full remaining concept set in a single pass, since deduplication requires global context. The resulting VIGOR vocabulary contains 3,011 concepts across the four concept levels.

\begin{figure}[ht]
\centering
\begin{tcolorbox}[colback=white,colframe=black!75!white,colbacktitle=black!75!white,width=\textwidth,title=Prompt for Filtering Candidate Concepts (Part 1)]
\begin{Verbatim}[fontsize=\scriptsize,breaklines=true,breakanywhere=true]
SYSTEM:
You are an expert in computer vision, visual perception, and ontology design. Your task is to evaluate visual concepts for inclusion in a minimal, comprehensive vocabulary that can describe any natural image through composition.

You apply rigorous criteria and return structured JSON. You never explain or hedge -- just evaluate and output the requested format.

USER:
# Task

Evaluate each of the candidate visual concepts listed below. These concepts were extracted from computer vision dataset annotations and are candidates for a minimal, comprehensive visual concept vocabulary. The vocabulary should contain only concepts that are atomic, visually grounded, and useful for describing natural images through composition.

---

# Criteria

## C1 -- Lexical well-formedness
Is this entry a clean, correctly spelled, standalone concept term?

REJECT if ANY of these apply:
- Misspelling or typo: "balck" (-> black), "brwon" (-> brown), "celing" (-> ceiling), "widshield" (-> windshield), "aentenna" (-> antenna).
- Sentence fragment: "is red", "are blue", "in black", "written in white", "color black", "colored green". A concept is a noun, adjective, or verb -- not a copula+adjective or preposition+adjective.
- Formatting artifact: "black+white", "red,white,blue", "red-and-white". Conjunctions with symbols are data artifacts, not concepts.
- Definition or synonym list: "bannister, banister, balustrade, balusters, handrail", "door, double door", "wiper (for windshield/screen)", "cap (container lid)". Parenthetical glosses and comma-separated alternatives are dictionary entries, not concept terms.
- Dangling phrase: "top of", "arm of a person", "back of chair", "side of building". If it requires a following noun to be meaningful, it is a fragment.

PASS if it is a real word or short phrase (1--3 words), correctly spelled, that can stand alone as a concept label.

## C2 -- Visual groundability
Can this concept be identified purely from pixel-level appearance in a photograph, without requiring external knowledge, text, audio, or reasoning about non-visible properties?

- PASS: "red", "furry", "sitting", "round", "kitchen" -- directly observable.
- FAIL: "expensive", "famous", "illegal", "tuesday", "owned", "imaginary" -- require inference beyond pixels.

Edge cases: "old" passes because wrinkles, patina, and wear are visible. "happy" passes because facial expression is visible. "dangerous" fails because it requires reasoning. "heavy" fails because weight is not visible. "frozen" passes because ice crystals and frost are visible surface properties.

## C3 -- Compositional atomicity
Is this concept a single, indivisible visual primitive, or is it a conjunction of two or more simpler concepts that each belong to a recognized visual category?

A concept is COMPOUND and should be rejected if it can be decomposed as:
- modifier + base where both parts are independently valid concepts:
  - color + color: "black and white" -> black + white
  - brightness + color: "dark blue" -> dark + blue, "bright red" -> bright + red, "pale pink" -> pale + pink
  - color + part: "red lips" -> red + lips, "blue eyes" -> blue + eyes, "white belly" -> white + belly, "yellow wheel" -> yellow + wheel
  - material + part: "metal handle" -> metal + handle, "wood beam" -> wood + beam, "glass pane" -> glass + pane
  - shape + part: "long sleeve" -> long + sleeve, "short mane" -> short + mane, "pointy ear" -> pointy + ear
  - state + part: "closed eye" -> closed + eye, "bent knees" -> bent + knees, "bare branch" -> bare + branch
  - object + part: "airplane door" -> airplane + door, "car hood" -> car + hood, "dog ear" -> dog + ear, "banana stem" -> banana + stem, "pizza crust" -> pizza + crust, "zebra's leg" -> zebra + leg
  - object + color: "painted green" -> painted + green, "blonde hair" -> blonde + hair
  - color + color compound: "reddish brown" -> red + brown, "greenish yellow" -> green + yellow, "blue-gray" -> blue + gray
\end{Verbatim}
\end{tcolorbox}
\caption{Prompt for filtering candidate concepts (part 1)}
\label{prompt:prompt_filter_concept_1}
\end{figure}

\begin{figure}[ht]
\centering
\begin{tcolorbox}[colback=white,colframe=black!75!white,colbacktitle=black!75!white,width=\textwidth,title=Prompt for Filtering Candidate Concepts (Part 2)]
\begin{Verbatim}[fontsize=\scriptsize,breaklines=true,breakanywhere=true]
A concept is ATOMIC and should be kept if:
- It names a single thing, property, or relationship: "red", "sleeve", "door", "rough", "sitting".
- Even if linguistically analyzable, the visual gestalt is distinct from its parts: "rainbow" is not just many colors side by side because it has an arc shape; "camouflage" is not just green+brown+pattern because it has a specific military-derived appearance.

The key test: can you express the same visual content by listing the components separately? If yes, it is compound. If something is lost, it is atomic.

## C4 -- Semantic level
Is the concept at the right level of abstraction?

- TOO_ABSTRACT: overly general, nearly every image contains it. Examples: "thing", "entity", "stuff", "area", "object", "item", "place".
- BASIC: the natural, default level humans use. Examples: "dog" not "mammal" or "golden retriever"; "red" not "color" or "crimson"; "running" not "moving" or "jogging at 6mph". This is the target level.
- TOO_SPECIFIC: fine-grained subtype not visually distinct from parent, or requires domain expertise. Examples: "Boeing 747-400", "Labrador retriever" instead of "dog", "Hepplewhite chair" instead of "chair", "terra cotta" instead of "orange-brown", "teal blue" instead of "teal".

## C5 -- Positive assertion
Does this concept describe the presence of a visual property, or the absence of one?

- POSITIVE and keep: "sleeve", "door", "striped", "bearded" -- describes something that IS visible.
- NEGATIVE and reject: "sleeveless", "doorless", "leafless", "headless", "strapless", "armless", "frameless", "unpeeled", "with no leaves" -- describes something that is NOT there.

Absence concepts are linguistically useful but are not visual primitives. You describe what you see, not what you do not see. The vocabulary can express "a shirt without sleeves" as "shirt" plus no "sleeve", rather than needing "sleeveless" as its own concept.

Exception: concepts where the "-less" form names a visually distinctive state that is not simply the parent minus the part: "wireless" names a technology category with its own visual signature; "seamless" describes a continuous surface, a positive visual property, not just the absence of seams. These are rare.

---

# Visual category assignment

Assign each concept to exactly ONE of these categories:

| Category | Definition | Examples |
|----------|------------|----------|
| color | Hue, saturation, brightness, color patterns | red, blue, golden, pale, dark, pastel, multicolored |
| edge | Boundary quality, contour properties, sharpness | sharp, blurry, smooth-edged, jagged, curved |
| texture | Surface pattern, tactile appearance, material finish | rough, smooth, woven, striped, dotted, furry, metallic |
| shape | Geometric form, spatial extent, proportions | round, rectangular, tall, thin, flat, elongated |
| part | Component of a larger object | wheel, handle, leg, roof, screen, wing, door |
| object | Discrete nameable entity or stuff region | dog, car, sky, grass, person, table, water |
| motion | Action, pose, dynamic state, body configuration | running, sitting, flying, falling, parked, open, closed |
| relation | Spatial, possessive, or interactive relationship | on, in, next to, holding, wearing, behind, above |

Disambiguation: "Furry" -> texture. "Wheel" -> part. "Standing" -> motion. "Wooden" -> texture. "Tall" -> shape. "Behind" -> relation.

---

# Concepts to evaluate
<concept list>

---
\end{Verbatim}
\end{tcolorbox}
\caption{Prompt for filtering candidate concepts (part 2)}
\label{prompt:prompt_filter_concept_2}
\end{figure}

\begin{figure}[ht]
\centering
\begin{tcolorbox}[colback=white,colframe=black!75!white,colbacktitle=black!75!white,width=\textwidth,title=Prompt for Filtering Candidate Concepts (Part 3)]
\begin{Verbatim}[fontsize=\scriptsize,breaklines=true,breakanywhere=true]
# Output format

Return a JSON array with exactly 50 objects, one per concept, in the same order. No text outside the JSON.

[
  {
    "id": 1,
    "concept": "<concept name>",
    "category": "<color|edge|texture|shape|part|object|motion|relation>",
    "c1_wellformed": true,
    "c2_groundable": true,
    "c3_atomic": true,
    "c3_decomposes_to": null,
    "c4_level": "basic",
    "c5_positive": true,
    "decision": "<keep|reject>",
    "reject_reasons": null
  }
]

Decision rules:
- "keep" if ALL: c1=true, c2=true, c3=true, c4=basic, c5=true.
- "reject" if ANY fails. List all failed criteria in reject_reasons, e.g., ["C1","C3"].

When c3_atomic is false, always populate c3_decomposes_to with the components, e.g., "dark + blue" or "airplane + door".

Be aggressive. When in doubt, reject.

Output ONLY the JSON array.
\end{Verbatim}
\end{tcolorbox}
\caption{Prompt for filtering candidate concepts (part 3)}
\label{prompt:prompt_filter_concept_3}
\end{figure}

\begin{figure}[ht]
\centering
\begin{tcolorbox}[colback=white,colframe=black!75!white,colbacktitle=black!75!white,width=\textwidth,title=Prompt for Cleaning Candidate Concepts (Part 1)]
\begin{Verbatim}[fontsize=\scriptsize,breaklines=true,breakanywhere=true]
SYSTEM:

You are an expert in computer vision, visual perception, and ontology design. You are performing the final cleaning pass on a visual concept vocabulary. The entire concept pool is provided below — your job is to identify and resolve redundancies, duplicates, and remaining quality issues that can only be detected when looking at the pool as a whole.

Return structured JSON. No explanation outside the JSON.


USER:

# Task

Below is the complete filtered visual concept pool, organized by visual category. Each category has already been through per-concept quality filtering (visual groundability, semantic level, atomicity, etc.). What remains are **pool-level** problems that require seeing multiple concepts together to detect:

- Spelling and hyphenation variants of the same concept
- Singular/plural duplicates
- Morphological variants (same root, different inflection)
- Near-synonyms (different words, visually indistinguishable)
- Concepts that decompose into other concepts IN THIS POOL
- Concepts that don't belong in their assigned category

Your job: produce a cleaned version of the pool by identifying every problematic cluster and resolving it.

---

# Rules

## R1 — Spelling / hyphenation variants
If the same concept appears with different spacing or hyphenation, keep exactly ONE canonical form (the most common English spelling, no hyphens when avoidable).

Examples: "band aid" / "band-aid" / "bandaid" → keep "bandaid". "cell phone" / "cellphone" → keep "cellphone". "t shirt" / "t-shirt" → keep "t-shirt".

## R2 — Singular / plural
Keep only the SINGULAR form. The vocabulary describes concept types, not quantities — plurality is a modifier, not a separate concept.

Examples: "cloud" / "clouds" → keep "cloud". "egg" / "eggs" → keep "egg". "biker" / "bikers" → keep "biker".

Exception: concepts that are inherently plural (the singular form is rarely used or means something different): "glasses" (eyewear), "scissors", "pants", "binoculars". Keep these as-is.

## R3 — Morphological variants
When multiple inflected forms of the same root appear, keep only the BASE form that best functions as a standalone concept label:

- For colors: keep the base adjective. "gray" / "graying" / "grayish" → keep "gray". "black" / "blackened" / "blackish" → keep "black". "yellow" / "yellowed" / "yellowish" → keep "yellow".
- For states: keep the form that names the observable state. "fade" / "faded" / "fading" → keep "faded" (the visible state). "shade" / "shaded" / "shading" / "shadowed" / "shadowing" / "shadowy" / "shady" → keep "shadowed" (the visible state in an image).
- For textures: keep the adjective form. "rust" / "rusted" / "rusty" → keep "rusty".

## R4 — Near-synonyms (visually indistinguishable)
If two concepts would produce indistinguishable photographs in most contexts, merge them. Keep the more common / general term.

Examples:
- "path" / "pathway" / "footpath" → keep "path"
- "shore" / "shoreline" / "coastline" / "coast" → keep "shore"
- "bathroom" / "restroom" / "washroom" → keep "bathroom"
- "couch" / "sofa" → keep "couch"
- "grey" / "gray" → keep "gray"
- "monochromatic" / "monochrome" → keep "monochrome"

Do NOT merge concepts that share a word but look different: "road" and "street" are both kept. "creek" and "river" are both kept.
\end{Verbatim}
\end{tcolorbox}
\caption{Prompt for cleaning candidate concepts (part 1)}
\label{prompt:prompt_clean_concept_1}
\end{figure}

\begin{figure}[ht]
\centering
\begin{tcolorbox}[colback=white,colframe=black!75!white,colbacktitle=black!75!white,width=\textwidth,title=Prompt for Cleaning Candidate Concepts (Part 2)]
\begin{Verbatim}[fontsize=\scriptsize,breaklines=true,breakanywhere=true]
## R5 — Reference-dependent composability
Now that you can see the full pool, check: is this concept expressible as a simple combination of 2-3 OTHER concepts that are IN THIS POOL?

The most important case: **person + role/activity compounds**. If both "person" and "running" are in the pool, then "runner" = person + running → remove "runner".

BUT keep role concepts that are visually distinctive through costume/equipment that ISN'T captured by the activity alone: "clown", "bride", "cowboy", "soldier", "firefighter".

Also check for object compounds: "alarm clock" = alarm + clock? No — "alarm" isn't the visual part. Keep it.

## R6 — Category misassignment
Flag concepts that are in the wrong category, or that don't cleanly belong to ANY category.

## R7 — Redundant age/gender person splits
Check the person-related concepts and remove aggregation terms like "people", "crowd", "group", "couple" if they are just person + quantity.

---

# The full concept pool

<concept pool>

---

# Output format

Return a JSON object with two sections:

```json
{
  "actions": [
    {
      "action": "merge",
      "rule": "R1",
      "cluster": ["band aid", "band-aid", "bandaid"],
      "keep": "bandaid",
      "remove": ["band aid", "band-aid"]
    },
    {
      "action": "remove",
      "rule": "R6",
      "concepts": ["backlit", "dim", "gloomy"],
      "reason": "lighting conditions, not colors"
    },
    {
      "action": "recategorize",
      "rule": "R6",
      "concept": "shadow",
      "from": "object",
      "to": "edge",
      "reason": "shadow is a visual phenomenon affecting edges/boundaries"
    }
  ],
  "summary": {
    "total_input": 0,
    "merges": 0,
    "removals": 0,
    "recategorizations": 0,
    "total_output": 0
  }
}
```

Be thorough. Scan every concept in every category. The goal is to reduce the pool to a truly minimal, non-redundant set where every concept earns its place and no two concepts are visually interchangeable.

Output ONLY the JSON object.
\end{Verbatim}
\end{tcolorbox}
\caption{Prompt for cleaning candidate concepts (part 2)}
\label{prompt:prompt_clean_concept_2}
\end{figure}

\subsection{Data Sourcing}
\label{app:data_sourcing}

To improve image-level concept coverage, we use \texttt{gpt-5.4} to inspect each image and identify visible concepts from the final vocabulary that are missing from the normalized source annotations. The model is constrained to select only concepts from the VIGOR vocabulary, so this step increases recall without introducing out-of-vocabulary labels. The prompt used for missing-concept identification is shown in Figure~\ref{prompt:prompt_identify_concept}.

\begin{figure}[ht]
\centering
\begin{tcolorbox}[colback=white,colframe=black!75!white,colbacktitle=black!75!white,width=\textwidth,title=Prompt for Identifying Missing Concepts from Images]
\begin{Verbatim}[fontsize=\scriptsize,breaklines=true,breakanywhere=true]
SYSTEM PROMPT:
You are a precise visual concept annotator for a vision research benchmark. You will be given an image and a set of visual concepts. Your task is to identify which concepts from the set are present in the image based on clear visual evidence. Be accurate and exhaustive - report every concept you can confidently identify, but never report concepts that lack clear visual support in the image. A concept must be visually grounded, not inferred from general knowledge about the world.

USER PROMPT:
Here is a set of visual concepts:
<concept set>

These concepts are already annotated for this image:
<annotated concepts>

Look at the provided image and identify which additional concepts from the full concept set are clearly present in the image but missing from the already annotated list. A concept is present only if there is clear visual evidence for it in the image. Do not report a concept based on assumption or world knowledge - only report it if you can directly see the evidence. Do not repeat concepts that are already annotated.

Respond ONLY with a JSON object in this exact format, no other text:

{"missing": ["concept_1", "concept_2", ...]}

The list should contain only concepts from the set above. Do not add any concepts outside the set. If none are missing, return {"missing": []}.
\end{Verbatim}
\end{tcolorbox}
\caption{Prompt for Identifying Missing Concepts from Images}
\label{prompt:prompt_identify_concept}
\end{figure}

\subsection{Annotator} 
\label{app:annotator}
The annotator stage is implemented as a common interface over heterogeneous proposal tools. Each tool receives the same RGB image and a text prompt derived from the target concept, and returns a floating-point mask at image resolution. The default prompt is the concept string itself. For color terms, the prompt is rewritten as ``\texttt{<color> object}'' so that proposal tools prefer visible color-bearing regions rather than treating the color word as an abstract attribute.
Continuous outputs are normalized and thresholded into binary masks. The default binary threshold is tool-specific: segmentation tools (SAM3, Grounded-SAM and CLIPSeg) use a lower threshold of 0.3, while CLIP CAM and MLLM attention ensemble use a higher threshold of 0.6 to suppress diffuse activation. 
The MLLM attention ensemble is implemented as the average of attention maps generated by multiple MLLMs (LLaVA-1.5-7B, LLaVA-OneVision-1.5-8B and GLM-4.6V-Flash in this work). The attention map of each individual MLLM is obtained by feeding ``\texttt{User: Describe the image. Assistant: A <concept>.}'' to the MLLM and collect the attention of the concept token to the image tokens, averaged across all layers and heads. 
When a concept has existing bounding box annotation, the candidate mask is post-multiplied by the box mask:
\[
M_{\mathrm{candidate}} \leftarrow M_{\mathrm{candidate}} \odot M_{\mathrm{bbox}}.
\]
This step preserves the shape proposed by the annotator but restricts it to the localized source region. For unlocalized concepts, no such spatial constraint is applied. We run all annotation tools on a server with 8 RTX A6000 GPUs, generating all annotations takes approximately 12 hours.

\subsection{Verifier} 
\label{app:verifier}
\begin{figure}[ht]
  \centering
  \includegraphics[width=\linewidth]{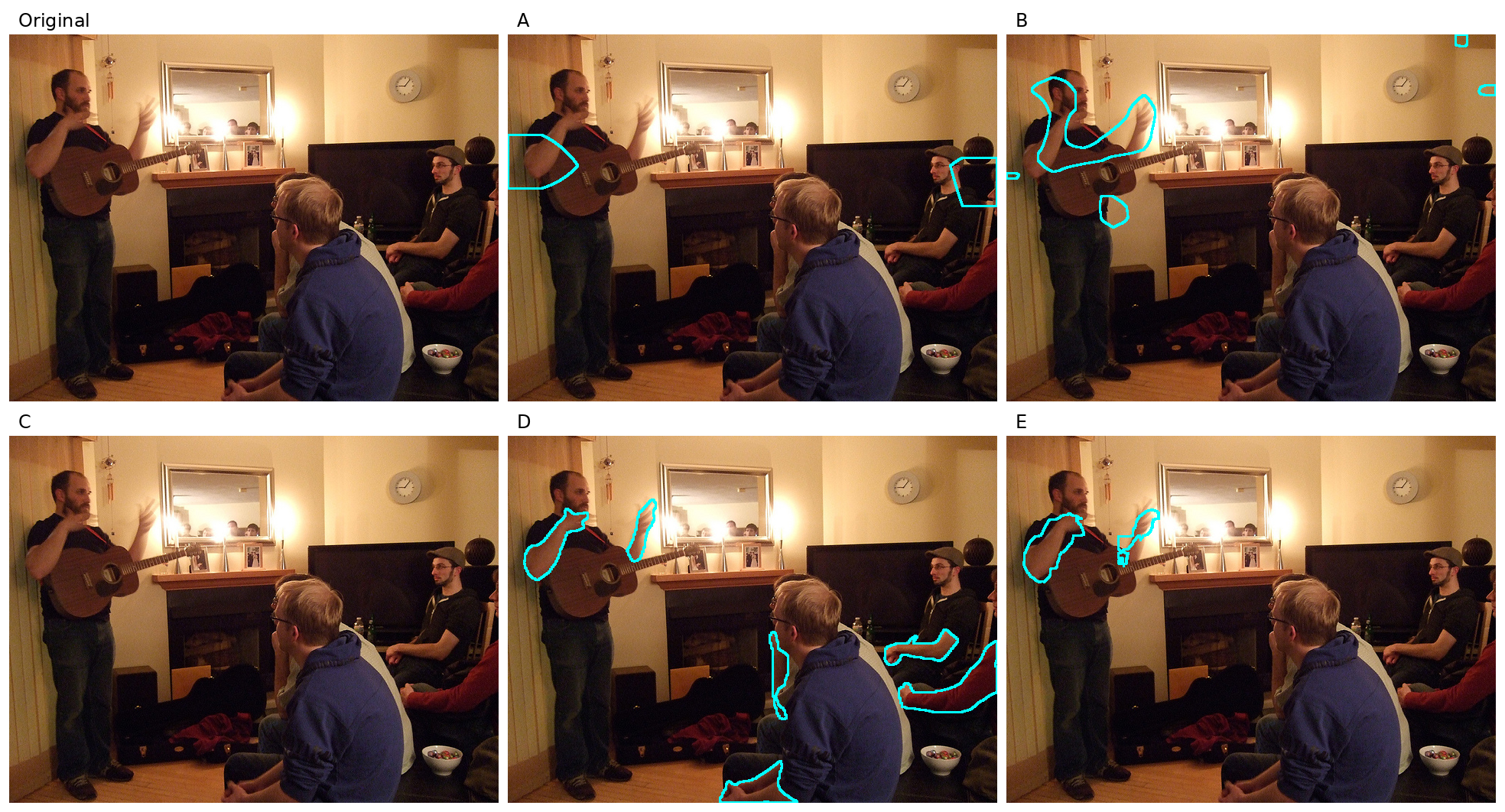}
  \caption{Example of composite panel of mask proposals.}
  \label{fig:proposal_example}
\end{figure}

The verifier input is a composite panel containing the original image and labeled candidate overlays. Each overlay displays the candidate mask contour in bright cyan on the same image, allowing the verifier to compare proposals under identical visual context. Candidate labels are mapped back to tool names only after the model returns a decision, preventing the prompt from favoring a specific annotator. The prompt used for the verification process is shown in Figure.~\ref{prompt:prompt_verification}. An example composite panel of mask proposals is shown in Figure.~\ref{fig:proposal_example}. We run all verifier models on a server with 8 RTX A6000 GPUs, verifying all annotations takes approximately 24 hours.

\begin{table}
\centering
\caption{Pairwise verifier agreement by concept level.}
\label{tab:verifier_agreement}
\begin{tabular}{lccccc}
\toprule
 & Overall & Primitive & Intermediate & Object & Scene \\
\midrule
Qwen / InternVL & 0.321 & 0.311 & 0.303 & 0.330 & 0.301 \\
Qwen / GLM & 0.445 & 0.467 & 0.488 & 0.424 & 0.481 \\
InternVL / GLM & 0.274 & 0.241 & 0.262 & 0.287 & 0.280 \\
\bottomrule
\end{tabular}
\end{table}
\begin{wraptable}{r}{0.36\textwidth}
\centering
\caption{Three-way agreement among MLLM verifiers by concept level. Agreement denotes the proportion of annotations for which all three verifiers selected the same decision.}
\label{tab:verifier_threeway_agreement}
\begin{tabular}{lc}
\toprule
& \textbf{Agreement} \\
\midrule
Overall & 0.183 \\
Primitive & 0.175 \\
Intermediate & 0.180 \\
Object & 0.184 \\
Scene & 0.201 \\
\bottomrule
\end{tabular}
\vspace{-10pt}
\end{wraptable}

We analyze verifier agreement over 35,748 candidate-mask decisions, including 7,346 primitive, 4,389 intermediate, 21,199 object, and 2,814 scene-level decisions.
Table~\ref{tab:verifier_agreement} reports pairwise verifier agreement by concept level. Agreement is highest between Qwen and GLM, with an overall agreement rate of $0.445$, and remains consistently high across primitive ($0.467$), intermediate ($0.488$), object ($0.424$), and scene ($0.481$) concepts. Qwen and InternVL show moderate agreement ($0.321$ overall), while InternVL and GLM agree less frequently ($0.274$ overall). These differences indicate that the verifiers are not redundant and provide complementary judgments over candidate masks.

Table~\ref{tab:verifier_threeway_agreement} reports the stricter rate at which all three verifiers select exactly the same decision. Exact three-way agreement is $0.183$ overall, with similar values for primitive ($0.175$), intermediate ($0.180$), object ($0.184$), and scene ($0.201$) concepts. The relatively low exact-agreement rate shows that candidate-mask selection is nontrivial, especially because the verifier must choose among multiple imperfect proposals or reject all candidates. This motivates our conservative majority-voting rule: a generated annotation is accepted only when at least two verifiers select the same candidate, reducing dependence on any single verifier while still allowing valid masks to be retained when full three-way consensus is unavailable.

\begin{figure}[ht]
\centering
\begin{tcolorbox}[colback=white,colframe=black!75!white,colbacktitle=black!75!white,width=\textwidth,title=Prompt for Mask Proposal Verification]
\begin{Verbatim}[fontsize=\scriptsize,breaklines=true,breakanywhere=true]
You will be given:
1. The original image.
2. A target concept.
3. 5 candidate annotation proposals, each shown as a visualization over the same image.
   - The proposal is surrounded by bright cyan contour.
   - Each proposal is only a candidate and may be wrong.
   - Candidate panels are labeled: A, B, C, D, E

Your task:
Choose exactly one of:
- A
- B
- C
- D
- E
- REJECT_ALL

Choose the proposal that best grounds the target concept in the image.
If none of the proposals is clearly valid, choose REJECT_ALL.

Important criteria:
1. Concept correctness:
   The selected proposal must correspond to the target concept, not just a visually salient or semantically related region.
   Example: for "red", highlight the red evidence, not the whole object unless the whole object is genuinely the evidence.
   Example: for "left of", the proposal must support the relation, not merely highlight one involved object.
2. Evidence precision:
   Prefer the proposal that marks the most relevant supporting evidence with minimal unnecessary area.
   Do not prefer overly large regions if a smaller region more precisely supports the concept.
3. Evidence sufficiency:
   The selected proposal must include enough visual evidence to justify the concept.
   Reject proposals that are too small, miss critical parts, or omit one side of a relation.
4. Faithfulness over plausibility:
   Do not reward a proposal just because it looks reasonable or because the concept is likely true in the image.
   Judge only whether the highlighted evidence actually supports the target concept.
5. Relation and comparison concepts:
   For relational concepts (such as "on", "behind", "holding", "next to"), the proposal should capture the necessary entities and the relation between them.
   For comparative concepts (such as "bigger", "brighter", "more reflective"), the proposal should include the evidence needed for the comparison.
6. Reject aggressively when needed:
   Choose REJECT_ALL if:
   - all proposals miss the concept,
   - all are too vague or too broad,
   - all are focused on the wrong object or wrong region,
   - the concept is not visually grounded by any proposal,
   - or every option is too weak to support a reasonable choice.

Decision rule:
- Select the label that is the best available grounding when it reasonably supports the concept, even if it is imperfect.
- Choose REJECT_ALL only when none of the candidates provides a reasonable visual grounding.

Output format:
Return a JSON object with exactly these fields:
{
  "decision": "<one of A, B, C, D, E, REJECT_ALL>",
  "confidence": 0.0,
  "reason": "brief explanation"
}

Requirements for the reason:
- Keep it under 60 words.
- State why the chosen proposal matches or why all should be rejected.
- Mention the key failure mode of the rejected proposals when relevant, such as:
  "too broad", "wrong concept", "misses critical evidence", "object-centric instead of concept-centric", "relation not grounded", or "insufficient visual support".

Prefer the best reasonable match over REJECT_ALL when one candidate is clearly more supportive than the rest.

Target concept: <concept>.
\end{Verbatim}
\end{tcolorbox}
\caption{Prompt for mask proposal verification}
\label{prompt:prompt_verification}
\end{figure}

\subsection{Human Evaluation}
\label{app:human_eval}

\definecolor{Reviewer1Header}{HTML}{CFE2F3}
\definecolor{Reviewer1Row}{HTML}{EEF6FC}
\definecolor{Reviewer2Header}{HTML}{D9EAD3}
\definecolor{Reviewer2Row}{HTML}{F1F8EE}
\definecolor{Reviewer3Header}{HTML}{FCE5CD}
\definecolor{Reviewer3Row}{HTML}{FFF3E8}
\definecolor{TableHeader}{HTML}{EAEAEA}

\begin{table}
\centering
\caption{Agreement proportions, precision, and recall by concept level for each reviewer.}
\label{tab:human_review_detail}
\small
\renewcommand{\arraystretch}{1.12}
\setlength{\tabcolsep}{4pt}

\begin{adjustbox}{max width=\textwidth}
\begin{tabular}{lccccccc}
\toprule
\rowcolor{TableHeader}
\textbf{Concept Level}
& \textbf{Strongly Agree}
& \textbf{Agree}
& \textbf{Neutral}
& \textbf{Disagree}
& \textbf{Strongly Disagree}
& \textbf{Precision}
& \textbf{Recall} \\
\midrule

\rowcolor{Reviewer1Header}
\multicolumn{8}{l}{\textbf{Reviewer 1}} \\
\rowcolor{Reviewer1Row}
Overall      & 0.748 & 0.062 & 0.062 & 0.096 & 0.032 & 0.798 & 0.833 \\
\rowcolor{Reviewer1Row}
Primitive    & 0.667 & 0.090 & 0.099 & 0.099 & 0.045 & 0.652 & 0.732 \\
\rowcolor{Reviewer1Row}
Intermediate & 0.774 & 0.000 & 0.038 & 0.132 & 0.057 & 0.696 & 0.762 \\
\rowcolor{Reviewer1Row}
Object       & 0.762 & 0.066 & 0.056 & 0.093 & 0.023 & 0.843 & 0.862 \\
\rowcolor{Reviewer1Row}
Scene        & 0.853 & 0.029 & 0.029 & 0.059 & 0.029 & 0.875 & 0.875 \\

\midrule
\rowcolor{Reviewer2Header}
\multicolumn{8}{l}{\textbf{Reviewer 2}} \\
\rowcolor{Reviewer2Row}
Overall      & 0.596 & 0.194 & 0.068 & 0.098 & 0.044 & 0.825 & 0.786 \\
\rowcolor{Reviewer2Row}
Primitive    & 0.459 & 0.252 & 0.153 & 0.117 & 0.018 & 0.674 & 0.646 \\
\rowcolor{Reviewer2Row}
Intermediate & 0.623 & 0.151 & 0.019 & 0.151 & 0.057 & 0.783 & 0.720 \\
\rowcolor{Reviewer2Row}
Object       & 0.642 & 0.189 & 0.043 & 0.086 & 0.040 & 0.882 & 0.840 \\
\rowcolor{Reviewer2Row}
Scene        & 0.588 & 0.118 & 0.088 & 0.059 & 0.147 & 0.688 & 0.688 \\

\midrule
\rowcolor{Reviewer3Header}
\multicolumn{8}{l}{\textbf{Reviewer 3}} \\
\rowcolor{Reviewer3Row}
Overall      & 0.272 & 0.606 & 0.012 & 0.108 & 0.002 & 0.909 & 0.866 \\
\rowcolor{Reviewer3Row}
Primitive    & 0.171 & 0.730 & 0.018 & 0.081 & 0.000 & 0.913 & 0.857 \\
\rowcolor{Reviewer3Row}
Intermediate & 0.226 & 0.585 & 0.019 & 0.151 & 0.019 & 0.826 & 0.760 \\
\rowcolor{Reviewer3Row}
Object       & 0.321 & 0.566 & 0.010 & 0.103 & 0.000 & 0.927 & 0.887 \\
\rowcolor{Reviewer3Row}
Scene        & 0.235 & 0.588 & 0.000 & 0.176 & 0.000 & 0.812 & 0.812 \\

\bottomrule
\end{tabular}
\end{adjustbox}
\end{table}

We provide additional details on the human evaluation of VIGOR annotations. We sample 500 annotations in total, including 111 primitive-level, 53 intermediate-level, 302 object-level, and 34 scene-level annotations. Each annotation is rated by three human reviewers on a five-point Likert scale: \textit{Strongly Agree}, \textit{Agree}, \textit{Neutral}, \textit{Disagree}, and \textit{Strongly Disagree}. We treat \textit{Strongly Agree} and \textit{Agree} as positive judgments, and treat \textit{Neutral}, \textit{Disagree}, and \textit{Strongly Disagree} as negative judgments.

Table~\ref{tab:human_review_detail} reports the rating distribution, precision, and recall for each reviewer and concept level. Overall, all three reviewers assign mostly positive ratings. Reviewer~1 gives $81.0\%$ positive ratings, Reviewer~2 gives $79.0\%$, and Reviewer~3 gives $87.8\%$. The reviewers differ in calibration: Reviewer~1 uses \textit{Strongly Agree} most frequently, Reviewer~3 uses \textit{Agree} most frequently, and Reviewer~2 is more balanced between the two positive labels. Despite these differences, the per-reviewer precision and recall remain consistently high. Overall precision is $0.798$, $0.825$, and $0.909$ for Reviewers~1--3, while overall recall is $0.833$, $0.786$, and $0.866$, respectively.

\begin{wraptable}{r}{0.36\textwidth}
\vspace{-10pt}
\centering
\caption{Inter-reviewer agreement by concept level. We treat ``Strongly Agree'' and ``Agree'' as positive, ``Strongly Disagree'', ``Disagree'', ``Neutral'' as negative, and calculate the agreement rate across all three human reviewers.}
\label{tab:human_agreement}
\begin{tabular}{lc}
\toprule
 & Agreement \\
\midrule
Overall & 0.778 \\
Primitive & 0.685 \\
Intermediate & 0.868 \\
Object & 0.801 \\
Scene & 0.735 \\
\bottomrule
\end{tabular}
\vspace{-10pt}
\end{wraptable}
The results also show systematic variation across concept levels. Object-level annotations receive the strongest validation, with high precision and recall across all reviewers. This is expected because object masks are typically spatially well-defined. Primitive and scene-level concepts are more variable: primitive concepts can correspond to diffuse visual properties such as colors or edges, while scene-level concepts often require broader contextual evidence. Intermediate concepts show high agreement but lower recall for some reviewers, reflecting the ambiguity of textures, materials, and shapes.

Table~\ref{tab:human_agreement} reports inter-reviewer agreement after binarizing the five-point ratings into positive and negative judgments. The overall agreement rate is $0.778$, indicating substantial consistency among reviewers. Agreement is highest for intermediate concepts ($0.868$), followed by object concepts ($0.801$), scene concepts ($0.735$), and primitive concepts ($0.685$). These results support the reliability of the automatic annotation pipeline while also showing that lower-level visual properties and scene-level evidence are inherently more ambiguous for human raters.

\noindent\textbf{Direction of verifier errors.} We further examine the cases in which the human judgment disagrees with the verifier panel's decision. Among these disagreements, $55.8\%$ are verifier false negatives, where the panel rejected a candidate mask that human reviewers judged to be valid, and $44.2\%$ are verifier false positives, where the panel retained a candidate that human reviewers judged to be incorrect. The higher proportion of false negatives indicates that the verifier panel is more likely to discard a valid candidate than to admit an incorrect one, which is the intended behavior of the conservative majority-voting rule: it trades some recall of valid masks for higher precision of the accepted annotations.

\section{Experiments} 

\subsection{Detailed Experimental Setup}
\label{app:detailed_setup}

A direct evaluation using model-generated captions would confound rationale correctness with caption coverage: different models may mention different subsets of the ground-truth concepts for the same image, making their rationale maps incomparable. To avoid this issue, we construct a fixed caption for each image in VIGOR using \texttt{gpt-5.4}. The caption is generated from the annotated concept set and is shared across all evaluated models. Each MLLM is then evaluated on the same image--caption pair, ensuring that differences in rationale scores reflect differences in visual grounding rather than differences in what concepts the model chooses to describe. The prompt used to construct captions from annotated concepts is shown in Figure~\ref{prompt:prompt_construct_caption}.

For each image $I_i$ and concept $c$, we identify the caption tokens corresponding to $c$. If a concept is tokenized into multiple subword tokens, we use all matched tokens; if a concept appears multiple times, we average over all occurrences. Let $\alpha^{(i)}_{\ell,h,t,p}$ denote the attention weight from caption token $t$ to image token $p$ at layer $\ell$ and head $h$. The concept-level rationale map is computed by averaging attention over matched concept tokens, layers, and heads:
\[
S_{i,c}(p)
=
\frac{1}{L H |\mathcal{T}_{i,c}|}
\sum_{\ell=1}^{L}
\sum_{h=1}^{H}
\sum_{t \in \mathcal{T}_{i,c}}
\alpha^{(i)}_{\ell,h,t,p},
\]
where $\mathcal{T}_{i,c}$ is the set of caption tokens matched to concept $c$, and $L$ and $H$ are the number of layers and attention heads. We reshape $S_{i,c}$ to the visual-token grid, upsample it to the original image resolution, and normalize it to $[0,1]$. The resulting dense map is compared against the ground-truth binary rationale mask $M_{i,c} \in \{0,1\}^{H_i \times W_i}$.

We report two metrics. First, \textit{Average Precision} (AP) measures how well the full rationale map ranks ground-truth rationale pixels above background pixels. Let $\Omega_i$ be the set of pixels in image $I_i$, and let $\pi$ be a permutation of pixels sorted by decreasing predicted score $S_{i,c}$. For the top $k$ ranked pixels, precision and recall are
\[
P_{i,c}(k)
=
\frac{\sum_{j=1}^{k} M_{i,c}(\pi_j)}{k},
\qquad
R_{i,c}(k)
=
\frac{\sum_{j=1}^{k} M_{i,c}(\pi_j)}
{\sum_{p \in \Omega_i} M_{i,c}(p)}.
\]
The AP for concept $c$ in image $I_i$ is then computed as the area under the precision--recall curve:
\[
\mathrm{AP}_{i,c}
=
\sum_{k=1}^{|\Omega_i|}
\bigl(R_{i,c}(k)-R_{i,c}(k-1)\bigr) P_{i,c}(k),
\]
with $R_{i,c}(0)=0$. AP rewards rationale maps that assign high scores to all pixels in the ground-truth evidence region.

Second, \textit{Pointing Accuracy} evaluates whether the most-attended pixel falls inside the ground-truth rationale mask:
\[
\mathrm{Point}_{i,c}
=
\mathbbm{1}
\left[
M_{i,c}
\left(
\arg\max_{p \in \Omega_i} S_{i,c}(p)
\right)
=1
\right].
\]
Unlike AP, Pointing Accuracy only evaluates the peak location of the rationale map. It therefore measures coarse localization, while AP evaluates the spatial alignment of the full attention distribution.

For a set of evaluated image--concept pairs $\mathcal{D}$, the final metrics are averaged as
\[
\mathrm{AP}
=
\frac{1}{|\mathcal{D}|}
\sum_{(i,c)\in\mathcal{D}}
\mathrm{AP}_{i,c},
\qquad
\mathrm{Pointing}
=
\frac{1}{|\mathcal{D}|}
\sum_{(i,c)\in\mathcal{D}}
\mathrm{Point}_{i,c}.
\]
We compute these scores both overall and separately for the four VIGOR concept levels: primitive, intermediate, object, and scene. We run all benchmark evaluations on a single RTX A6000 GPU, evaluating a single model takes approximately 10 minutes.

\begin{figure}[ht]
\centering
\begin{tcolorbox}[colback=white,colframe=black!75!white,colbacktitle=black!75!white,width=\textwidth,title=Prompt for Constructing Captions from Annotated Concepts]
\begin{Verbatim}[fontsize=\scriptsize,breaklines=true,breakanywhere=true]
SYSTEM:
You are a vision-language captioning assistant for an image annotation dataset.
Your job is to write captions that stay faithful to the actual visible image content. 
Use the provided concept list as guidance, but do not invent details that are not visually supported by the image.


USER:
Here is the list of annotated concepts for this image:
<concept list>

Task:
- Look at the image.
- Write a detailed natural-language caption for the image.
- Use the concept annotations as required content to cover in the caption.
- Include all annotated concepts in the caption.
- Keep the caption faithful to the visible image content.
- Do not add details that are not visible in the image.

Respond ONLY with a JSON object in this exact format:

{"caption": "your caption here"}
\end{Verbatim}
\end{tcolorbox}
\caption{Prompt for constructing captions from annotated concepts}
\label{prompt:prompt_construct_caption}
\end{figure}

\subsection{Validation of the Fixed Captions}
\label{app:caption_validation}
\begin{table}[t]
\centering
\small
\caption{
Human validation of the fixed VIGOR captions. Three evaluators independently rated 500 randomly sampled \texttt{gpt-5.4} captions for faithfulness to the image and annotated concepts, and for fluency, on a five-point Likert scale. We report the mean rating $\pm$ standard error.
}
\label{tab:caption_validation}
\begin{tabular}{lcc}
\toprule
Evaluator & Faithfulness & Fluency \\
\midrule
Evaluator A & 4.86 $\pm$ 0.04 & 4.79 $\pm$ 0.06 \\
Evaluator B & 4.91 $\pm$ 0.06 & 4.85 $\pm$ 0.04 \\
Evaluator C & 4.82 $\pm$ 0.07 & 4.82 $\pm$ 0.05 \\
\midrule
Average     & 4.86 & 4.82 \\
\bottomrule
\end{tabular}
\end{table}

\noindent\textbf{Human validation.} Because all models are evaluated on captions constructed by \texttt{gpt-5.4}, we verify that these captions faithfully express the annotated concepts. We randomly sample 500 captions and ask three evaluators to independently rate each caption on a five-point Likert scale along two dimensions: \emph{faithfulness}, whether the caption accurately describes the image and its annotated concepts without introducing unsupported content, and \emph{fluency}, whether the caption is grammatical and coherent. Table~\ref{tab:caption_validation} reports the results. The captions receive an average score of $4.86$ for faithfulness and $4.82$ for fluency, and the three evaluators are closely aligned, indicating that the fixed captions accurately express the annotated visual concepts in natural language.

\noindent\textbf{Perplexity analysis.} A potential confound of the fixed-caption protocol is distribution shift: a model might obtain lower rationale scores simply because the \texttt{gpt-5.4} captions differ from the text it would generate itself. To test this, we compute each evaluated model's perplexity on the fixed captions and correlate it with the model's overall AP. The Pearson correlation is weak and not statistically significant ($r=0.328$, $p=0.102$), indicating that differences in caption likelihood are not a major driver of rationale correctness in our evaluation. Sec.~\ref{app:generated_caption} further shows that model rankings remain largely consistent when models generate their own captions.

\subsection{Evaluation with Model-Generated Captions}
\label{app:generated_caption}
\begin{table}[t]
\centering
\footnotesize
\caption{
Evaluation with model-generated captions. Output correctness (CIDEr, ROUGE-L) and rationale correctness (AP, Pointing Accuracy) are computed on each model's own caption; fixed-caption scores are from Table~\ref{tab:main_results}.
}
\label{tab:generated_caption}
\setlength{\tabcolsep}{6pt}
\renewcommand{\arraystretch}{0.9}
\begin{tabular}{lcccccc}
\toprule
\multirow{2}{*}{Model} & \multicolumn{2}{c}{Output correctness} & \multicolumn{2}{c}{Generated caption} & \multicolumn{2}{c}{Fixed caption} \\
\cmidrule(lr){2-3} \cmidrule(lr){4-5} \cmidrule(lr){6-7}
 & CIDEr & ROUGE-L & AP & Pointing & AP & Pointing \\
\midrule
Qwen3-VL-2B    & 0.494 & 0.229 & 34.2 & 42.0 & 25.2 & 31.9 \\
Qwen3-VL-4B    & 0.466 & 0.216 & 37.0 & 44.0 & 28.3 & 34.5 \\
Qwen3-VL-8B    & 0.468 & 0.217 & 34.2 & 40.4 & 26.3 & 32.2 \\
InternVL3.5-2B & 0.613 & 0.257 & 40.6 & 50.3 & 28.8 & 36.5 \\
InternVL3.5-4B & 0.551 & 0.249 & 41.8 & 52.1 & 29.7 & 37.2 \\
InternVL3.5-8B & 0.542 & 0.242 & 42.9 & 53.2 & 30.6 & 38.2 \\
Gemma-3-4B     & 0.341 & 0.202 & 39.7 & 50.7 & 26.1 & 34.2 \\
Gemma-3-12B    & 0.369 & 0.209 & 40.0 & 49.8 & 27.8 & 35.2 \\
\midrule
Average        & 0.481 & 0.228 & 38.8 & 47.8 & 27.9 & 35.0 \\
\bottomrule
\end{tabular}
\vspace{-8pt}
\end{table}

\noindent\textbf{Protocol.} The fixed-caption protocol evaluates all models on identical image--concept pairs. As an end-to-end complement that more directly tests whether MLLMs ``look at what they say,'' we let each model caption every VIGOR image itself and score both components of double correctness on that output. Each model is prompted to describe the image, and its generated caption is used both for output scoring and for rationale extraction. Output correctness is computed with CIDEr~\cite{vedantam2015cider} and ROUGE-L~\cite{lin2004rouge} using the COCO caption evaluation toolkit~\cite{chen2015microsoft}. For rationale correctness, we match tokens in the generated caption to concepts in the image's VIGOR annotation set and extract the concept-token rationale map exactly as in the fixed-caption protocol; generated concepts without a VIGOR annotation are not scored, and annotated concepts that the model does not mention are excluded for that model. Because each model is thus evaluated on its own self-selected concept subset, absolute scores are not directly comparable to the fixed-caption protocol, and we focus on rank consistency between the two settings.

\noindent\textbf{Results.} Table~\ref{tab:generated_caption} reports eight models from three families. First, generated captions yield higher rationale scores: averaged over the eight models, AP rises from $27.9$ to $38.8$ and Pointing Accuracy from $35.0$ to $47.8$, since models tend to mention concepts they can readily recognize and ground, whereas the fixed-caption protocol holds the concept set constant. Second, model rankings are largely preserved across the two protocols, with Spearman $\rho=0.862$ for AP and $\rho=0.833$ for Pointing Accuracy, so the comparative conclusions of Sec.~\ref{sec:main_results} are robust to the choice of protocol. Third, output and rationale correctness are only weakly related: averaging the two output metrics and the two rationale metrics per model gives Pearson $r=0.176$. A model that writes a better caption does not necessarily ground the concepts it mentions in more appropriate evidence, which supports treating rationale correctness as a complementary axis of evaluation rather than a by-product of output quality.

\subsection{Benchmark Results on Other Rationale Extraction Methods}
\label{app:additional_results}
We report the benchmark results using GradCAM in Table.~\ref{tab:gradcam} and LogitLens in Table.~\ref{tab:logitlens}.

\begin{table}
\caption{
Benchmark results of GradCAM on VIGOR benchmark. We report AP and Pointing Accuracy for overall grounding and for primitive, intermediate, object, and scene-level concepts. The best performing model in each column is \textbf{bolded} and the second best in \underline{underlined}.
}
\label{tab:gradcam}
\centering
\resizebox{0.9\textwidth}{!}{
\begin{tabular}{lcccccccccc}
\toprule
\multirow{2}{*}{Model} & \multicolumn{2}{c}{Overall} & \multicolumn{2}{c}{Primitive} & \multicolumn{2}{c}{Intermediate} & \multicolumn{2}{c}{Object} & \multicolumn{2}{c}{Scene} \\
\cmidrule(lr){2-3} \cmidrule(lr){4-5} \cmidrule(lr){6-7} \cmidrule(lr){8-9} \cmidrule(lr){10-11}
 & AP & Pointing & AP & Pointing & mAP & Pointing & mAP & Pointing & mAP & Pointing \\
\midrule
Random & 5.8 & 5.7 & 4.0 & 3.9 & 1.9 & 1.8 & 5.0 & 4.9 & 17.7 & 17.3 \\
\midrule
InternVL3-1B & 12.8 & 17.4 & 10.1 & 16.0 & 6.4 & 9.0 & 13.9 & 20.6 & 23.3 & 23.7 \\
\midrule
InternVL3-2B & 12.9 & 18.4 & 12.1 & 17.9 & 5.8 & 7.5 & 14.2 & 21.2 & 23.2 & 31.0 \\
InternVL3.5-2B & 17.4 & 20.5 & 15.5 & 19.9 & 8.8 & 11.7 & 21.2 & 25.7 & 22.5 & 20.8 \\
Qwen2-VL-2B & 16.0 & 17.3 & 15.8 & 18.5 & 8.7 & 10.0 & 18.5 & 20.2 & 22.4 & 21.7 \\
Qwen3-VL-2B & 19.9 & 24.7 & 21.4 & 31.3 & 9.1 & 11.7 & 23.8 & 29.8 & 27.2 & 29.8 \\
\midrule
Ministral3-3B & 13.2 & 18.4 & 15.8 & 24.5 & 5.2 & 8.0 & 14.3 & 21.5 & 23.6 & 24.4 \\
Qwen2.5-VL-3B & 14.7 & 16.1 & 11.9 & 13.4 & 6.5 & 7.8 & 18.4 & 20.0 & 20.5 & 21.4 \\
Gemma3-4B & 11.3 & 13.8 & 11.9 & 16.1 & 4.7 & 6.0 & 12.7 & 17.0 & 18.8 & 17.1 \\
InternVL3.5-4B & 16.6 & 20.2 & 15.9 & 20.4 & 7.4 & 10.0 & 20.0 & 24.2 & 24.0 & 26.7 \\
Qwen3-VL-4B & 17.2 & 21.9 & 18.4 & 24.4 & 8.0 & 10.8 & 20.2 & 26.7 & 24.8 & 26.8 \\
\midrule
Qwen2-VL-7B & 19.1 & 21.5 & 16.9 & 19.5 & 9.6 & 11.5 & 23.4 & 26.2 & 24.8 & 27.4 \\
Qwen2.5-VL-7B & 11.8 & 9.2 & 10.3 & 8.1 & 4.8 & 3.1 & 14.0 & 10.2 & 19.5 & 18.3 \\
InternVL3-8B & 15.6 & 20.8 & 15.0 & 20.6 & 6.9 & 9.2 & 18.1 & 25.7 & 24.6 & 28.0 \\
InternVL3.5-8B & 16.3 & 19.6 & 16.0 & 19.8 & 7.4 & 9.8 & 19.2 & 23.8 & 23.2 & 24.9 \\
Ministral3-8B & 13.6 & 19.1 & 17.1 & 26.3 & 5.5 & 8.0 & 14.9 & 22.1 & 23.2 & 26.7 \\
Qwen3-VL-8B & 15.8 & 20.1 & 17.0 & 21.5 & 7.2 & 9.8 & 18.9 & 25.1 & 21.6 & 23.4 \\
\midrule
Gemma3-12B & 13.3 & 18.8 & 18.6 & 26.9 & 6.0 & 9.1 & 14.9 & 22.4 & 19.1 & 20.8 \\
InternVL3-14B & 14.3 & 17.9 & 15.8 & 19.0 & 6.7 & 8.9 & 15.8 & 20.6 & 23.1 & 25.0 \\
InternVL3.5-14B & 17.1 & 22.2 & 16.4 & 20.9 & 7.9 & 11.2 & 21.0 & 27.6 & 22.3 & 26.1 \\
Ministral3-14B & 14.4 & 20.5 & 17.4 & 24.3 & 5.5 & 8.3 & 15.8 & 24.2 & 25.3 & 30.2 \\
\midrule
Gemma3-27B & 10.1 & 12.2 & 11.3 & 17.3 & 4.1 & 5.3 & 11.0 & 15.1 & 17.9 & 13.2 \\
Qwen2.5-VL-32B & 16.9 & 21.9 & 18.8 & 22.4 & 7.5 & 10.3 & 20.1 & 26.4 & 23.6 & 29.4 \\
Qwen3-VL-32B & 14.9 & 20.9 & 16.7 & 23.4 & 6.7 & 10.7 & 17.5 & 26.3 & 21.3 & 22.2 \\
\bottomrule
\end{tabular}
}
\end{table}

\begin{table}
\caption{
Benchmark results of LogitLens on VIGOR benchmark. We report AP and Pointing Accuracy for overall grounding and for primitive, intermediate, object, and scene-level concepts. The best performing model in each column is \textbf{bolded} and the second best in \underline{underlined}.
}
\label{tab:logitlens}
\centering
\resizebox{0.9\textwidth}{!}{
\begin{tabular}{lcccccccccc}
\toprule
\multirow{2}{*}{Model} & \multicolumn{2}{c}{Overall} & \multicolumn{2}{c}{Primitive} & \multicolumn{2}{c}{Intermediate} & \multicolumn{2}{c}{Object} & \multicolumn{2}{c}{Scene} \\
\cmidrule(lr){2-3} \cmidrule(lr){4-5} \cmidrule(lr){6-7} \cmidrule(lr){8-9} \cmidrule(lr){10-11}
 & AP & Pointing & AP & Pointing & mAP & Pointing & mAP & Pointing & mAP & Pointing \\
\midrule
Random & 5.8 & 5.7 & 4.0 & 3.9 & 1.9 & 1.8 & 5.0 & 4.9 & 17.7 & 17.3 \\
\midrule
InternVL3-1B & 11.7 & 13.6 & 9.2 & 10.2 & 5.8 & 6.8 & 13.1 & 16.1 & 20.2 & 21.1 \\
InternVL3.5-1B & 10.5 & 12.0 & 10.4 & 12.2 & 4.4 & 5.5 & 11.3 & 13.8 & 19.9 & 18.2 \\
\midrule
InternVL3-2B & 13.8 & 17.6 & 16.1 & 23.0 & 6.7 & 8.9 & 15.6 & 21.3 & 20.5 & 20.1 \\
InternVL3.5-2B & 13.0 & 15.1 & 13.3 & 17.7 & 6.7 & 8.0 & 14.8 & 18.0 & 19.4 & 18.1 \\
Qwen2-VL-2B & 15.5 & 17.6 & 16.0 & 19.9 & 9.2 & 11.5 & 17.5 & 20.3 & 21.1 & 20.0 \\
Qwen3-VL-2B & 13.8 & 16.9 & 10.9 & 12.1 & 7.4 & 9.3 & 16.4 & 21.4 & 19.5 & 20.0 \\
\midrule
Ministral3-3B & 12.3 & 16.0 & 16.2 & 22.7 & 5.8 & 7.9 & 13.1 & 18.1 & 20.3 & 21.5 \\
Qwen2.5-VL-3B & 13.4 & 15.9 & 12.9 & 16.0 & 6.9 & 9.0 & 15.6 & 19.2 & 19.7 & 19.5 \\
Gemma3-4B & 14.6 & 17.5 & 15.0 & 18.4 & 6.0 & 7.2 & 17.4 & 22.0 & 22.4 & 23.0 \\
InternVL3.5-4B & 14.5 & 16.6 & 17.0 & 20.9 & 6.7 & 7.8 & 16.2 & 19.9 & 22.6 & 20.9 \\
Qwen3-VL-4B & 17.0 & 21.1 & 19.3 & 24.2 & 9.2 & 11.7 & 19.8 & 25.9 & 21.7 & 22.0 \\
\midrule
Qwen2-VL-7B & 13.8 & 15.5 & 12.7 & 14.0 & 7.8 & 10.0 & 15.8 & 17.7 & 20.0 & 20.8 \\
Qwen2.5-VL-7B & 17.0 & 19.6 & 16.8 & 19.5 & 9.7 & 12.3 & 20.1 & 23.4 & 21.6 & 23.0 \\
InternVL3-8B & 13.4 & 16.1 & 13.7 & 17.2 & 6.7 & 8.1 & 15.6 & 20.3 & 19.6 & 17.9 \\
InternVL3.5-8B & 13.9 & 16.5 & 14.2 & 17.3 & 6.9 & 8.2 & 16.2 & 20.2 & 20.1 & 20.3 \\
Ministral3-8B & 12.4 & 15.6 & 16.7 & 22.6 & 5.5 & 7.1 & 13.3 & 17.3 & 20.9 & 23.0 \\
Qwen3-VL-8B & 13.2 & 16.3 & 16.5 & 23.4 & 7.1 & 8.5 & 14.5 & 18.9 & 19.3 & 18.7 \\
\midrule
Gemma3-12B & 16.1 & 21.2 & 19.7 & 26.9 & 7.2 & 10.1 & 19.0 & 25.5 & 22.4 & 25.1 \\
InternVL3-14B & 15.7 & 18.4 & 16.9 & 21.7 & 7.9 & 9.1 & 18.6 & 22.6 & 21.0 & 21.1 \\
InternVL3.5-14B & 14.7 & 17.4 & 14.8 & 18.0 & 7.1 & 9.1 & 17.3 & 21.4 & 21.0 & 20.4 \\
Ministral3-14B & 14.1 & 17.7 & 19.5 & 25.7 & 6.3 & 8.1 & 15.7 & 20.2 & 21.0 & 23.3 \\
\midrule
Gemma3-27B & 16.2 & 21.9 & 20.6 & 29.2 & 7.4 & 10.6 & 18.8 & 26.0 & 22.7 & 26.6 \\
Qwen2.5-VL-32B & 13.9 & 17.8 & 15.9 & 19.9 & 6.7 & 9.3 & 16.2 & 21.6 & 19.6 & 21.4 \\
Qwen3-VL-32B & 12.8 & 16.3 & 15.1 & 20.3 & 6.4 & 9.2 & 14.0 & 18.6 & 20.4 & 20.8 \\
InternVL3-38B & 13.4 & 16.4 & 17.3 & 22.1 & 6.5 & 8.5 & 15.0 & 19.7 & 19.9 & 18.9 \\
InternVL3.5-38B & 13.8 & 16.7 & 14.0 & 16.9 & 7.5 & 9.3 & 15.5 & 19.7 & 21.1 & 21.5 \\
\bottomrule
\end{tabular}
}
\end{table}


\clearpage
\newpage
\section*{NeurIPS Paper Checklist}

\begin{enumerate}

\item {\bf Claims}
    \item[] Question: Do the main claims made in the abstract and introduction accurately reflect the paper's contributions and scope?
    \item[] Answer: \answerYes{} 
    \item[] Justification: The main claims made in the abstract and introduction accurately reflect the main paper's contributions and scope.
    \item[] Guidelines:
    \begin{itemize}
        \item The answer \answerNA{} means that the abstract and introduction do not include the claims made in the paper.
        \item The abstract and/or introduction should clearly state the claims made, including the contributions made in the paper and important assumptions and limitations. A \answerNo{} or \answerNA{} answer to this question will not be perceived well by the reviewers. 
        \item The claims made should match theoretical and experimental results, and reflect how much the results can be expected to generalize to other settings. 
        \item It is fine to include aspirational goals as motivation as long as it is clear that these goals are not attained by the paper. 
    \end{itemize}

\item {\bf Limitations}
    \item[] Question: Does the paper discuss the limitations of the work performed by the authors?
    \item[] Answer: \answerYes{} 
    \item[] Justification: We include a through limitation discussion in \S Appendix \ref{app:limitations}.
    \item[] Guidelines:
    \begin{itemize}
        \item The answer \answerNA{} means that the paper has no limitation while the answer \answerNo{} means that the paper has limitations, but those are not discussed in the paper. 
        \item The authors are encouraged to create a separate ``Limitations'' section in their paper.
        \item The paper should point out any strong assumptions and how robust the results are to violations of these assumptions (e.g., independence assumptions, noiseless settings, model well-specification, asymptotic approximations only holding locally). The authors should reflect on how these assumptions might be violated in practice and what the implications would be.
        \item The authors should reflect on the scope of the claims made, e.g., if the approach was only tested on a few datasets or with a few runs. In general, empirical results often depend on implicit assumptions, which should be articulated.
        \item The authors should reflect on the factors that influence the performance of the approach. For example, a facial recognition algorithm may perform poorly when image resolution is low or images are taken in low lighting. Or a speech-to-text system might not be used reliably to provide closed captions for online lectures because it fails to handle technical jargon.
        \item The authors should discuss the computational efficiency of the proposed algorithms and how they scale with dataset size.
        \item If applicable, the authors should discuss possible limitations of their approach to address problems of privacy and fairness.
        \item While the authors might fear that complete honesty about limitations might be used by reviewers as grounds for rejection, a worse outcome might be that reviewers discover limitations that aren't acknowledged in the paper. The authors should use their best judgment and recognize that individual actions in favor of transparency play an important role in developing norms that preserve the integrity of the community. Reviewers will be specifically instructed to not penalize honesty concerning limitations.
    \end{itemize}

\item {\bf Theory assumptions and proofs}
    \item[] Question: For each theoretical result, does the paper provide the full set of assumptions and a complete (and correct) proof?
    \item[] Answer: \answerNA{} 
    \item[] Justification: There's no theoretical results in this paper.
    \item[] Guidelines:
    \begin{itemize}
        \item The answer \answerNA{} means that the paper does not include theoretical results. 
        \item All the theorems, formulas, and proofs in the paper should be numbered and cross-referenced.
        \item All assumptions should be clearly stated or referenced in the statement of any theorems.
        \item The proofs can either appear in the main paper or the supplemental material, but if they appear in the supplemental material, the authors are encouraged to provide a short proof sketch to provide intuition. 
        \item Inversely, any informal proof provided in the core of the paper should be complemented by formal proofs provided in appendix or supplemental material.
        \item Theorems and Lemmas that the proof relies upon should be properly referenced. 
    \end{itemize}

    \item {\bf Experimental result reproducibility}
    \item[] Question: Does the paper fully disclose all the information needed to reproduce the main experimental results of the paper to the extent that it affects the main claims and/or conclusions of the paper (regardless of whether the code and data are provided or not)?
    \item[] Answer: \answerYes{} 
    \item[] Justification: The paper fully disclose all the information needed to reproduce the main experimental results of the paper. \S Appendix \ref{app:concept_vocab}--\ref{app:verifier}, and \ref{app:detailed_setup} contain all necessary details for reproducing our results.
    \item[] Guidelines:
    \begin{itemize}
        \item The answer \answerNA{} means that the paper does not include experiments.
        \item If the paper includes experiments, a \answerNo{} answer to this question will not be perceived well by the reviewers: Making the paper reproducible is important, regardless of whether the code and data are provided or not.
        \item If the contribution is a dataset and\slash or model, the authors should describe the steps taken to make their results reproducible or verifiable. 
        \item Depending on the contribution, reproducibility can be accomplished in various ways. For example, if the contribution is a novel architecture, describing the architecture fully might suffice, or if the contribution is a specific model and empirical evaluation, it may be necessary to either make it possible for others to replicate the model with the same dataset, or provide access to the model. In general. releasing code and data is often one good way to accomplish this, but reproducibility can also be provided via detailed instructions for how to replicate the results, access to a hosted model (e.g., in the case of a large language model), releasing of a model checkpoint, or other means that are appropriate to the research performed.
        \item While NeurIPS does not require releasing code, the conference does require all submissions to provide some reasonable avenue for reproducibility, which may depend on the nature of the contribution. For example
        \begin{enumerate}
            \item If the contribution is primarily a new algorithm, the paper should make it clear how to reproduce that algorithm.
            \item If the contribution is primarily a new model architecture, the paper should describe the architecture clearly and fully.
            \item If the contribution is a new model (e.g., a large language model), then there should either be a way to access this model for reproducing the results or a way to reproduce the model (e.g., with an open-source dataset or instructions for how to construct the dataset).
            \item We recognize that reproducibility may be tricky in some cases, in which case authors are welcome to describe the particular way they provide for reproducibility. In the case of closed-source models, it may be that access to the model is limited in some way (e.g., to registered users), but it should be possible for other researchers to have some path to reproducing or verifying the results.
        \end{enumerate}
    \end{itemize}

\item {\bf Open access to data and code}
    \item[] Question: Does the paper provide open access to the data and code, with sufficient instructions to faithfully reproduce the main experimental results, as described in supplemental material?
    \item[] Answer: \answerYes{} 
    \item[] Justification: This paper provides open access to the data and code, and include instructions for running the code.
    \item[] Guidelines:
    \begin{itemize}
        \item The answer \answerNA{} means that paper does not include experiments requiring code.
        \item Please see the NeurIPS code and data submission guidelines (\url{https://neurips.cc/public/guides/CodeSubmissionPolicy}) for more details.
        \item While we encourage the release of code and data, we understand that this might not be possible, so \answerNo{} is an acceptable answer. Papers cannot be rejected simply for not including code, unless this is central to the contribution (e.g., for a new open-source benchmark).
        \item The instructions should contain the exact command and environment needed to run to reproduce the results. See the NeurIPS code and data submission guidelines (\url{https://neurips.cc/public/guides/CodeSubmissionPolicy}) for more details.
        \item The authors should provide instructions on data access and preparation, including how to access the raw data, preprocessed data, intermediate data, and generated data, etc.
        \item The authors should provide scripts to reproduce all experimental results for the new proposed method and baselines. If only a subset of experiments are reproducible, they should state which ones are omitted from the script and why.
        \item At submission time, to preserve anonymity, the authors should release anonymized versions (if applicable).
        \item Providing as much information as possible in supplemental material (appended to the paper) is recommended, but including URLs to data and code is permitted.
    \end{itemize}

\item {\bf Experimental setting/details}
    \item[] Question: Does the paper specify all the training and test details (e.g., data splits, hyperparameters, how they were chosen, type of optimizer) necessary to understand the results?
    \item[] Answer: \answerYes{} 
    \item[] Justification: \S Appendix \ref{app:detailed_setup} contain all experimental details.
    \item[] Guidelines:
    \begin{itemize}
        \item The answer \answerNA{} means that the paper does not include experiments.
        \item The experimental setting should be presented in the core of the paper to a level of detail that is necessary to appreciate the results and make sense of them.
        \item The full details can be provided either with the code, in appendix, or as supplemental material.
    \end{itemize}

\item {\bf Experiment statistical significance}
    \item[] Question: Does the paper report error bars suitably and correctly defined or other appropriate information about the statistical significance of the experiments?
    \item[] Answer: \answerYes{} 
    \item[] Justification: We provide statistical significance analyses of the Pearson correlation in \S Appendix~\ref{sec:hallucination}.
    \item[] Guidelines:
    \begin{itemize}
        \item The answer \answerNA{} means that the paper does not include experiments.
        \item The authors should answer \answerYes{} if the results are accompanied by error bars, confidence intervals, or statistical significance tests, at least for the experiments that support the main claims of the paper.
        \item The factors of variability that the error bars are capturing should be clearly stated (for example, train/test split, initialization, random drawing of some parameter, or overall run with given experimental conditions).
        \item The method for calculating the error bars should be explained (closed form formula, call to a library function, bootstrap, etc.)
        \item The assumptions made should be given (e.g., Normally distributed errors).
        \item It should be clear whether the error bar is the standard deviation or the standard error of the mean.
        \item It is OK to report 1-sigma error bars, but one should state it. The authors should preferably report a 2-sigma error bar than state that they have a 96\% CI, if the hypothesis of Normality of errors is not verified.
        \item For asymmetric distributions, the authors should be careful not to show in tables or figures symmetric error bars that would yield results that are out of range (e.g., negative error rates).
        \item If error bars are reported in tables or plots, the authors should explain in the text how they were calculated and reference the corresponding figures or tables in the text.
    \end{itemize}

\item {\bf Experiments compute resources}
    \item[] Question: For each experiment, does the paper provide sufficient information on the computer resources (type of compute workers, memory, time of execution) needed to reproduce the experiments?
    \item[] Answer: \answerYes{} 
    \item[] Justification: We include all experiments compute resources in the Appendix.
    \item[] Guidelines:
    \begin{itemize}
        \item The answer \answerNA{} means that the paper does not include experiments.
        \item The paper should indicate the type of compute workers CPU or GPU, internal cluster, or cloud provider, including relevant memory and storage.
        \item The paper should provide the amount of compute required for each of the individual experimental runs as well as estimate the total compute. 
        \item The paper should disclose whether the full research project required more compute than the experiments reported in the paper (e.g., preliminary or failed experiments that didn't make it into the paper). 
    \end{itemize}
    
\item {\bf Code of ethics}
    \item[] Question: Does the research conducted in the paper conform, in every respect, with the NeurIPS Code of Ethics \url{https://neurips.cc/public/EthicsGuidelines}?
    \item[] Answer: \answerYes{} 
    \item[] Justification: We confirm the research conducted in the paper conform, in every respect, with the NeurIPS Code of Ethics.
    \item[] Guidelines:
    \begin{itemize}
        \item The answer \answerNA{} means that the authors have not reviewed the NeurIPS Code of Ethics.
        \item If the authors answer \answerNo, they should explain the special circumstances that require a deviation from the Code of Ethics.
        \item The authors should make sure to preserve anonymity (e.g., if there is a special consideration due to laws or regulations in their jurisdiction).
    \end{itemize}

\item {\bf Broader impacts}
    \item[] Question: Does the paper discuss both potential positive societal impacts and negative societal impacts of the work performed?
    \item[] Answer: \answerYes{} 
    \item[] Justification: The paper discusses potential positive societal impacts in \S Appendix~\ref{app:broader_impact}.
    \item[] Guidelines:
    \begin{itemize}
        \item The answer \answerNA{} means that there is no societal impact of the work performed.
        \item If the authors answer \answerNA{} or \answerNo, they should explain why their work has no societal impact or why the paper does not address societal impact.
        \item Examples of negative societal impacts include potential malicious or unintended uses (e.g., disinformation, generating fake profiles, surveillance), fairness considerations (e.g., deployment of technologies that could make decisions that unfairly impact specific groups), privacy considerations, and security considerations.
        \item The conference expects that many papers will be foundational research and not tied to particular applications, let alone deployments. However, if there is a direct path to any negative applications, the authors should point it out. For example, it is legitimate to point out that an improvement in the quality of generative models could be used to generate Deepfakes for disinformation. On the other hand, it is not needed to point out that a generic algorithm for optimizing neural networks could enable people to train models that generate Deepfakes faster.
        \item The authors should consider possible harms that could arise when the technology is being used as intended and functioning correctly, harms that could arise when the technology is being used as intended but gives incorrect results, and harms following from (intentional or unintentional) misuse of the technology.
        \item If there are negative societal impacts, the authors could also discuss possible mitigation strategies (e.g., gated release of models, providing defenses in addition to attacks, mechanisms for monitoring misuse, mechanisms to monitor how a system learns from feedback over time, improving the efficiency and accessibility of ML).
    \end{itemize}
    
\item {\bf Safeguards}
    \item[] Question: Does the paper describe safeguards that have been put in place for responsible release of data or models that have a high risk for misuse (e.g., pre-trained language models, image generators, or scraped datasets)?
    \item[] Answer: \answerNA{} 
    \item[] Justification: The paper poses no such risks.
    \item[] Guidelines:
    \begin{itemize}
        \item The answer \answerNA{} means that the paper poses no such risks.
        \item Released models that have a high risk for misuse or dual-use should be released with necessary safeguards to allow for controlled use of the model, for example by requiring that users adhere to usage guidelines or restrictions to access the model or implementing safety filters. 
        \item Datasets that have been scraped from the Internet could pose safety risks. The authors should describe how they avoided releasing unsafe images.
        \item We recognize that providing effective safeguards is challenging, and many papers do not require this, but we encourage authors to take this into account and make a best faith effort.
    \end{itemize}

\item {\bf Licenses for existing assets}
    \item[] Question: Are the creators or original owners of assets (e.g., code, data, models), used in the paper, properly credited and are the license and terms of use explicitly mentioned and properly respected?
    \item[] Answer: \answerYes{} 
    \item[] Justification: The creators or original owners of assets used in the paper are properly credited and are respected for the license and terms of use explicitly mentioned. 
    \item[] Guidelines:
    \begin{itemize}
        \item The answer \answerNA{} means that the paper does not use existing assets.
        \item The authors should cite the original paper that produced the code package or dataset.
        \item The authors should state which version of the asset is used and, if possible, include a URL.
        \item The name of the license (e.g., CC-BY 4.0) should be included for each asset.
        \item For scraped data from a particular source (e.g., website), the copyright and terms of service of that source should be provided.
        \item If assets are released, the license, copyright information, and terms of use in the package should be provided. For popular datasets, \url{paperswithcode.com/datasets} has curated licenses for some datasets. Their licensing guide can help determine the license of a dataset.
        \item For existing datasets that are re-packaged, both the original license and the license of the derived asset (if it has changed) should be provided.
        \item If this information is not available online, the authors are encouraged to reach out to the asset's creators.
    \end{itemize}

\item {\bf New assets}
    \item[] Question: Are new assets introduced in the paper well documented and is the documentation provided alongside the assets?
    \item[] Answer: \answerYes{} 
    \item[] Justification: We document all assets.
    \item[] Guidelines:
    \begin{itemize}
        \item The answer \answerNA{} means that the paper does not release new assets.
        \item Researchers should communicate the details of the dataset\slash code\slash model as part of their submissions via structured templates. This includes details about training, license, limitations, etc. 
        \item The paper should discuss whether and how consent was obtained from people whose asset is used.
        \item At submission time, remember to anonymize your assets (if applicable). You can either create an anonymized URL or include an anonymized zip file.
    \end{itemize}

\item {\bf Crowdsourcing and research with human subjects}
    \item[] Question: For crowdsourcing experiments and research with human subjects, does the paper include the full text of instructions given to participants and screenshots, if applicable, as well as details about compensation (if any)? 
    \item[] Answer: \answerNA{} 
    \item[] Justification: The paper does not involve crowdsourcing nor research with human subjects.
    \item[] Guidelines:
    \begin{itemize}
        \item The answer \answerNA{} means that the paper does not involve crowdsourcing nor research with human subjects.
        \item Including this information in the supplemental material is fine, but if the main contribution of the paper involves human subjects, then as much detail as possible should be included in the main paper. 
        \item According to the NeurIPS Code of Ethics, workers involved in data collection, curation, or other labor should be paid at least the minimum wage in the country of the data collector. 
    \end{itemize}

\item {\bf Institutional review board (IRB) approvals or equivalent for research with human subjects}
    \item[] Question: Does the paper describe potential risks incurred by study participants, whether such risks were disclosed to the subjects, and whether Institutional Review Board (IRB) approvals (or an equivalent approval/review based on the requirements of your country or institution) were obtained?
    \item[] Answer: \answerNA{} 
    \item[] Justification: The paper does not involve crowdsourcing nor research with human subjects.
    \item[] Guidelines:
    \begin{itemize}
        \item The answer \answerNA{} means that the paper does not involve crowdsourcing nor research with human subjects.
        \item Depending on the country in which research is conducted, IRB approval (or equivalent) may be required for any human subjects research. If you obtained IRB approval, you should clearly state this in the paper. 
        \item We recognize that the procedures for this may vary significantly between institutions and locations, and we expect authors to adhere to the NeurIPS Code of Ethics and the guidelines for their institution. 
        \item For initial submissions, do not include any information that would break anonymity (if applicable), such as the institution conducting the review.
    \end{itemize}

\item {\bf Declaration of LLM usage}
    \item[] Question: Does the paper describe the usage of LLMs if it is an important, original, or non-standard component of the core methods in this research? Note that if the LLM is used only for writing, editing, or formatting purposes and does \emph{not} impact the core methodology, scientific rigor, or originality of the research, declaration is not required.
    \item[] Answer: \answerYes{} 
    \item[] Justification: We describe how LLMs/MLLMs are being used as part of the tools for constructing the benchmark.
    \item[] Guidelines:
    \begin{itemize}
        \item The answer \answerNA{} means that the core method development in this research does not involve LLMs as any important, original, or non-standard components.
        \item Please refer to our LLM policy in the NeurIPS handbook for what should or should not be described.
    \end{itemize}

\end{enumerate}

\end{document}